\documentclass[11pt]{article}

\usepackage[final]{acl}

\usepackage{times}
\usepackage{latexsym}

\usepackage[T1]{fontenc}
\usepackage[utf8]{inputenc}
\usepackage{microtype}
\usepackage{csquotes}
\usepackage{graphicx}
\usepackage{booktabs}
\usepackage[normalem]{ulem}

\usepackage{amsmath}
\usepackage{amssymb}

\usepackage{multirow}
\usepackage{makecell}
\usepackage{listings}
\lstdefinestyle{promptstyle}{
  basicstyle=\ttfamily\footnotesize,
  breaklines=true,
  breakatwhitespace=false,
  columns=fullflexible,
  keepspaces=true,
  upquote=true,
  showstringspaces=false,
  frame=single,
  framerule=0.3pt,
  xleftmargin=0.5em,
  xrightmargin=0.5em,
  aboveskip=0.75em,
  belowskip=0.75em,
  captionpos=t,
}
\usepackage[shortlabels]{enumitem}
\setlist{nosep, leftmargin=1.4em}

\newif\ifcomments
\commentstrue



\ifcomments

\newcommand{\cut}[1]{}
\newcommand{\maybecut}[1]{\textcolor{orange}{#1}}
\newcommand{\debating}[1]{\textcolor{red}{#1}}

\newcommand{\remove}[1]{\textcolor{green}{#1}}
\newcommand{\removehide}[1]{}
\newcommand{\alt}[1]{\textcolor{brown}{#1}}
\newcommand{\SA}[1]{\textcolor{red}{({\bf SA:} #1)}}
\newcommand{\SB}[1]{\textcolor{blue}{({\bf SB:} #1)}}
\newcommand{\JT}[1]{\textcolor{violet}{({\bf JT:} #1)}}

\else
    \newcommand{\SA}[1]{}
    \newcommand{\SB}[1]{}
    \newcommand{\JT}[1]{}
    \newcommand{\sp}[1]{}
    \newcommand{\maybecut}[1]{}
    \newcommand{\debating}[1]{}
    
    \newcommand{\remove}[1]{}
    \newcommand{\removehide}[1]{}
    \newcommand{\alt}[1]{}

\fi

\newcommand{\avgattn}{Avg.\ Attention}

\title{Tokengeist: Multi-Turn Attribution Tracing in Agentic Conversations}

\author{
 \textbf{Jessica Tang\textsuperscript{1}\thanks{Work done during internship at Microsoft Research.}},
 \textbf{Shraddha Barke\textsuperscript{2}},
 \textbf{Sharad Agarwal\textsuperscript{2}},
\\
 \textsuperscript{1}Brown University,
 \textsuperscript{2}Microsoft Research,
\\
 \small{
   \textbf{Correspondence:} jessica\_tang1@brown.edu, sbarke@microsoft.com, sagarwal@microsoft.com
 }
}

\begin{document}
\maketitle
\begin{abstract}
When a model answers after several turns of dialogue and tool use, an overseer
needs to know which earlier turns and intermediate steps it rests
on. Attribution is the natural primitive for answering this, but existing
context attribution methods process the full transcript in a single pass: they
recover surface-level dependencies while missing the layered structure of 
multi-turn traces, in which an assistant turn is itself a derived artifact that
condenses or transforms earlier context.
We introduce \textit{multi-turn context attribution} (MTCA): given a target span
in a model response, the task of tracing attribution backward across turns to
identify not only which prior turns were directly relevant, but also how those
turns themselves depended on earlier context. We propose \textsc{Tokengeist}, an
attribution-method-agnostic and scalable framework that recovers full dependency
paths by casting attribution as a recursive traversal of a directed acyclic
graph over conversation turns. We will release \textsc{MTCABench}, a
benchmark of 3{,}970 target spans across 685 multi-turn conversations,
annotated with gold provenance graphs reaching depths of up to 17, across four dependency types.
Across four open-weight models, flat attribution methods fail to recover multi-hop dependencies, 
with source recall improving from 11--38\% under flat attribution to 82--94\% under \textsc{Tokengeist}. 
Our results reveal a systematic failure mode of single-pass attribution which we term
\textit{provenance collapse}.
\end{abstract}

\section{Introduction}


Large language models (LLMs) are increasingly deployed as agents that act over long horizons: calling tools, reading and restating results, and committing to consequential actions several turns later.
Multi-turn dialogues, tool-calling agents, and retrieval-augmented pipelines all produce contexts in which a model's final answer may depend on information introduced many turns prior---possibly after that information has been summarized, reformatted, or transformed by intermediate assistant turns.
The stakes are highest in agentic conversations, where tools are automatically invoked for the user in high-value scenarios such as finance, shopping, and enterprise workflows, and where checking a single span by hand would mean re-examining several prior turns.
This raises a concrete question for oversight, debugging, and trust: \emph{given a specific claim in a model's response, what chain of prior context explains its provenance?}

Context attribution is the natural primitive for answering this, and prior work has studied it primarily in single-turn settings.
Methods include attention-based heuristics \citep{cohen2025learning}, gradient-based approaches \citep{yin2022interpreting}, learned surrogate models \citep{wang2025attntrace,cohen2024contextcite}, and prompting the model to explain its own reasoning \citep{tahaei-etal-2024-efficient, laban-etal-2024-summary}.
However, these methods typically treat the available context as a \emph{flat} sequence: a prompt or retrieved context followed by a response.
When applied to multi-turn conversations, they can assign importance scores to earlier tokens or spans, but they do not explicitly recover the \emph{layered} structure of dialogue.
 
We argue this gap is consequential. We hypothesize that attribution in
multi-turn settings has a \emph{non-linear, multi-hop structure}: the answer to
turn $N$ may depend on a claim in turn $N{-}4$, which in turn depended on a tool
output in turn $3$. A flat attribution method looking at the full context window
can only identify which turns \emph{appear} relevant, not how relevance was \emph{propagated} forward. An early source's content is absorbed by the turns that follow it, and positional degradation over long contexts~\citep{liu-etal-2024-lost, laban2026llms} compounds this challenge (Appendix~\ref{sec:appendix-signal-depth}).

\begin{figure}[t]
  \centering
  \includegraphics[width=0.9\linewidth]{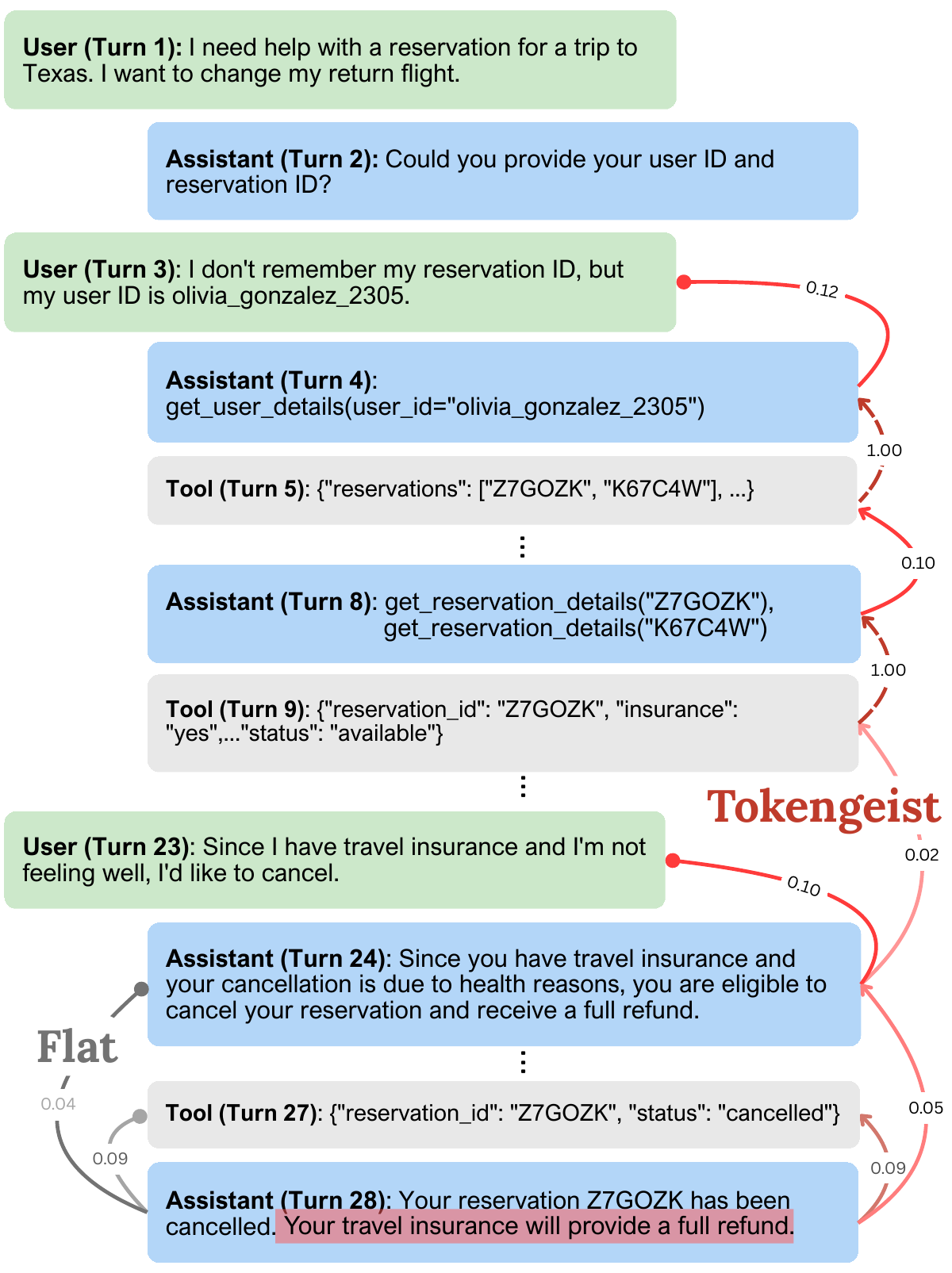}
  \caption{%
    \textbf{Recursive provenance tracing example.}
    Right-hand arrows (red) show \textsc{Tokengeist}'s recursive trace, stopping at user-authored turns (T23, T3).
    Left-hand arrows (gray) show the flat attribution trace, stopping at assistant-authored turn T27.
    Labels are raw attribution scores; edges are retained following Eq.~\ref{eq:rel-threshold}.}
  \label{fig:motivating_example}
\end{figure}

\paragraph{Motivating example.}
Figure~\ref{fig:motivating_example} illustrates this challenge on an agentic airline customer-support conversation.
The agent's claim \textit{``your travel insurance will provide a full refund''} (T28) traces back through an intermediate eligibility determination (T24) that combined two earlier sources: the user's stated health reason (T23) and an insurance field in a tool result (T9) fetched via a lookup chain seeded by the user ID six turns prior (T3).
\emph{Flat} attribution applied to T28 assigns its highest score to T27---the cancellation tool result confirming the booking was cancelled but says nothing about insurance eligibility---and misses the sources T23 and T3.
A reader handed T27 as the explanation would be inspecting a turn that cannot support or refute the insurance claim at all.

\paragraph{Multi-Turn Context Attribution (MTCA).}
We formalize this problem as MTCA.
Given a target span in the final model response, the goal is to recover its full provenance: which prior turns or context does it depend on, and through what chain of intermediate steps did that information reach the response?
Unlike flat attribution, MTCA must account for the layered structure of dialogues.

\paragraph{Tokengeist.}
We propose \textsc{Tokengeist}, a framework for MTCA that represents provenance as a DAG over conversation turns and is agnostic to the choice of base attribution method.
Starting from the target span, \textsc{Tokengeist} applies a flat attribution method to identify the most relevant prior turns, then recurses into each candidate intermediate turn until it reaches original user input sources.
It reads only the recorded transcript, so it can trace a conversation produced by any agent, including one served behind a closed API; when the attributing model is itself the generator, the attention it reads \emph{is} the attention that produced the text.
 
\paragraph{MTCABench.}
To evaluate MTCA, we introduce \textsc{MTCABench}, a benchmark of 3{,}970 annotated target spans across 685 multi-turn conversations 
(from ConFETTI~\citep{alkhouli2025confetti} and TauBench~\citep{yao2024tau}) 
with gold provenance graphs, which we will release. Conversations include tool-calling traces and cases that stress-test LLM failure modes including hallucination and distracting turns.
 
\paragraph{Results.}
Recursion improves on the same attribution method applied flat for every method, model and dataset we test; with the two strongest attention-based methods it gains 23--44 Edge F1 points and raises source recall on TauBench from 18\% to 87\% on average.
Flat attribution stops at the nearest assistant turn, while \textsc{Tokengeist} reaches the originating user or tool event.
Controls isolate recursion, rather than the base attributor or the structural edge, as the source of the gain (Section~\ref{sec:experiments}).


\section{Related Work}
\label{sec:related-work}

\paragraph{Single-turn context attribution.}
A growing body of work identifies which parts of an input influence a model's
generated output. Early token- or feature-level methods include contrastive
explanations~\citep{yin2022interpreting}, which identify the tokens explaining
why a model preferred one prediction over another, and perturbation-based
approaches such as LIME~\citep{ribeiro2016should} and
SHAP~\citep{lundberg2017unified}.

More recent work connects generated statements to evidence in the provided
context. ContextCite~\citep{cohen2024contextcite} learns a surrogate model that
predicts how a generation changes when context sources are included or
excluded, while AT2~\citep{cohen2025learning} avoids expensive ablations by
combining attention-head signals under ablation-derived supervision.
Long-context traceback methods such as TracLLM~\citep{wang2025tracllm} and
AttnTrace~\citep{wang2025attntrace} locate sentences, passages, or paragraphs
in long inputs via informed search, contribution-score denoising, or
attention-based traceback.

Several benchmarks evaluate evidence grounding:
\citet{li2024attributionbench} show that automatic attribution
evaluation remains difficult; \citet{li2024towards} target
knowledge-aware generation over structured sources; and
\citet{hirsch2025laquer} localize user-selected generated spans to
source spans. A related line evaluates attributed generation and citation
quality~\citep{gao2023enabling, yu2026c2}.
All of these treat the context as a flat input and return relevant tokens,
spans, paragraphs, or documents for a single generation.
MTCA instead asks for the recursive provenance structure of a conversation:
not only which event influenced the target span, but which \textit{earlier}
events influenced that event.

\paragraph{Multi-turn and long-context understanding.}
\citet{liu-etal-2024-lost} and \citet{laban2026llms} document that LLMs do
not uniformly use information distributed across long contexts.
\citet{bai-etal-2024-mt} and \citet{kwan-etal-2024-mt}
evaluate multi-turn conversational abilities, and recent tool-use
benchmarks~\citep{alkhouli2025confetti,yao2024tau,patil2025berkeley} move
evaluation closer to realistic histories where actions depend on prior turns,
tool calls, and tool results. All score task success, response quality, or
tool-call correctness rather than whether a generated span can be explained by
recovering its multi-hop provenance graph.

\paragraph{Graph-structured provenance.}
Our formulation represents provenance as a DAG whose edges
point from a later event to the earlier event it depends on, connecting MTCA
to workflow provenance and program-dependence analysis, which capture how
entities are derived from prior entities and
activities~\citep{acar2010graph} or make data and control dependencies
explicit in program executions~\citep{ferrante1987program}.
Recent NLP work has begun to study attribution over structured artifacts;
\citet{suri-etal-2025-follow}, for example, ground model responses in
flowchart components via graph-based reasoning. However in MTCA, the
graph is not given to the model as an input artifact but must be derived from a
conversation trace whose intermediate nodes may themselves be model-generated.
This makes recursive tracing central, since a method may identify a nearby
assistant summary while missing the original source from which it was
derived---a failure mode we formalize as \emph{provenance collapse}.

\section{Problem Formulation}
\label{sec:problem-formulation}



\paragraph{Notation.}
A \textit{conversation} $\mathcal{C}$ is a sequence of turns $T_1, T_2, \ldots, T_N$, where each turn 
$T_i = (r_i, c_i)$ 
consists of a \textit{role} $r_i \in \{\textsc{System}, \textsc{User}, \textsc{Assistant}, \textsc{Tool}\}$ and \textit{content} $c_i$, a sequence of tokens.
We refer to turns with $r_i \in \{\textsc{System}, \textsc{User}\}$ as \textit{exogenous turns}, since their content is externally supplied.
Turns with $r_i = \textsc{Assistant}$ are \textit{endogenous turns}, generated by the model.
Turns with $r_i = \textsc{Tool}$ are \textit{tool-result turns}: their content is produced by an external tool executor rather than the model, so they carry no model-generated attention weights and are handled separately during attribution (Section~\ref{sec:tokengeist}).
$T_T$ denotes the \textit{target turn}: any assistant turn in $\mathcal{C}$ whose content we aim to explain. A \textit{target span} is a contiguous sub-sequence $\bar{y} = (y_a, \ldots, y_b)$ of $T_T$'s content, with $1 \le a \le b \le |c_T|$.

\subsection{Single-Turn Attribution}

Attribution operates across three granularities.

\textbf{Token level} is where attribution is computed.
Let $\mathcal{A}$ denote a \textit{base attribution method} (Section~\ref{sec:backends}): given a target span $\bar{y}$ (a contiguous slice of output tokens) and a context $X$ (the preceding token sequence), $\mathcal{A}$ scores every source token $x_j \in X$ by how much it influenced $\bar{y}$.

\textbf{Sentence level} is the scoring unit.
We segment $X$ into sentences $\mathrm{seg}(X) = (s_1, \ldots, s_M)$ and score each as an atomic unit over its token index range, which reduces noise from attention spikes on punctuation and common tokens:
\begin{equation}
  \mathbf{a} = \mathcal{A}(\bar{y},\, s_1,\ldots,s_M) \in \mathbb{R}_{\ge 0}^{M}.
  \label{eq:span-scores}
\end{equation}

\textbf{Turn level} is the provenance unit.
When $X$ concatenates multiple turns each $s_i$ belongs to exactly one turn, and turn aggregation takes the strongest sentence-level signal within each:
\begin{equation}
  \hat{a}_k = \max_{\,i\,:\,s_i \in T_k} a_i,
  \label{eq:turn-agg}
\end{equation}
yielding one score $\hat{a}_k \ge 0$ per preceding turn $T_k$.

In \textit{flat attribution}, $\mathcal{A}$ is applied once with $X$ equal to all turns preceding the target, then turn aggregation (Eq.~\ref{eq:turn-agg}) yields one score per turn.

\subsection{Multi-Turn Context Attribution (MTCA)}

Given $\mathcal{C}$, $\bar{y}$, and $\mathcal{A}$, the goal of MTCA is to recover a \textit{provenance graph} $G = (V, E, w)$: a weighted DAG over conversation turns in which $T_T$ is the unique root, each edge $(T_i, T_j) \in E$ denotes that $T_i$ depends on $T_j$ with $\mathrm{idx}(j) < \mathrm{idx}(i)$, and leaf nodes are exogenous turns.
The DAG accommodates multi-source dependencies, where a single turn draws on multiple independent earlier turns.





\section{Tokengeist}
\label{sec:tokengeist}
 
\begin{figure*}[t]
  \centering
  \includegraphics[width=0.85\textwidth]{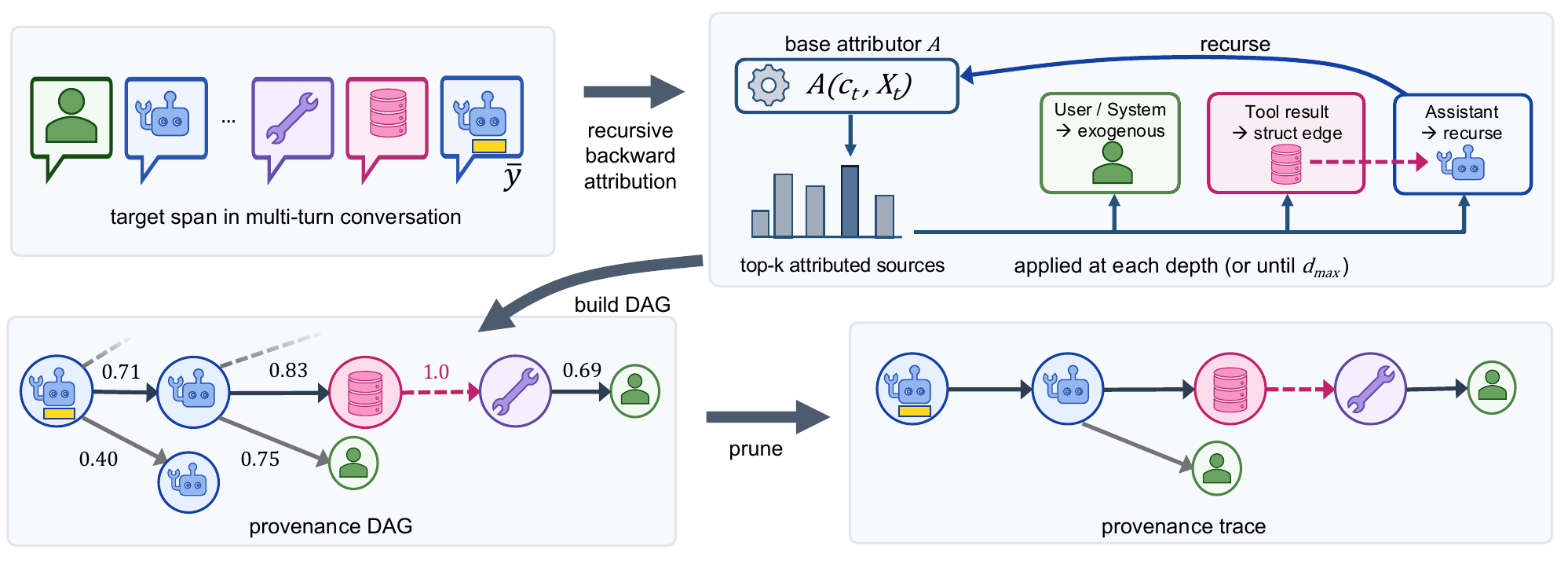}  
  \caption{\textbf{Overview of \textsc{Tokengeist}.}
  From a target span $\bar{y}$, a base attributor $\mathcal{A}$ is applied recursively to the top-$k$ prior turns at each step.
  Role-aware dispatch sends user / system turns to exogenous leaves, tool-result turns along a deterministic structural edge (dashed) to their invoking tool-call turn, and assistant turns back into the recursion.
  The resulting DAG is pruned by relative-threshold extraction (Section~\ref{sec:tokengeist}).}
  \label{fig:method}
\end{figure*}

\subsection{Recursive attribution}

\textsc{Tokengeist} (Figure~\ref{fig:method}; Appendix~\ref{sec:appendix-procedure}) builds $G$ by backward breadth-first search (BFS) from $T_T$, using $\bar{y}$ as the initial attribution target.
At each frontier node $v$ representing turn $T_t$, the tracer:
(i)~concatenates turns $T_0,\ldots,T_{t-1}$ into a context string $X_t$;
(ii)~calls $\mathcal{A}(c_t,\, \mathrm{seg}(X_t))$ to obtain sentence-level scores over $X_t$, where $c_t$ is the token span being explained: $\bar{y}$ at depth~0, or the full token sequence of turn $T_t$ at deeper steps;
(iii)~applies turn aggregation (Eq.~\ref{eq:turn-agg}) to convert sentence scores to one score per preceding turn;
(iv)~discards turns with aggregated score~$< \theta$ (default $\theta\!=\!0$) and selects the top~$k$ turns (default $k\!=\!3$).

The role of each selected turn determines how recursion proceeds.
An \textbf{exogenous turn} is added as a leaf node and recursion stops.
A \textbf{tool-result turn} carries no model-generated attention weights, so \textsc{Tokengeist} instead follows a deterministic \emph{structural edge} (score~$1.0$) to the nearest preceding assistant turn that issued the invocation and resumes recursive attribution there, encoding the causal chain \textbf{target} $\to$ \textbf{tool result} $\xrightarrow{\textit{struct.}}$ \textbf{tool invocation} $\to$ \textbf{argument sources}.
An \textbf{assistant turn} is added as a node and recursed into, attributing its full content against the earlier context.
Recursion also terminates once the maximum depth $d_{\max}$ is reached (default $d_{\max} = 8$).

Flat attribution is the special case $d_{\max}\!=\!1$: $\mathcal{A}$ is applied once to $\bar{y}$ over the full context, without recursion.
It therefore cannot recover intermediate dependencies when the turns it attributes are themselves endogenous turns that summarize or restate earlier information---a failure mode we formalize as \textbf{provenance collapse}, in which the true exogenous source receives low attribution because the model never attended to it directly in the final turn, having already incorporated its content into an intermediate endogenous turn.

\subsection{Design choices}

\paragraph{Turn-level DAG nodes.}
Sentence-level scores from $\mathcal{A}$ locate evidence, but the output of each recursive step is the prior turn to recurse into.
This matches dialogue representations that treat utterances as basic semantic
units~\citep{zhang-etal-2023-dialog}; in our setting, a turn is the corresponding
utterance-level unit, i.e., one contribution by a single agent.
Turn-level nodes therefore give recursion a stable semantic boundary.
Span F1 (Section~\ref{sec:metrics}) measures the precision of within-turn span
localization.

\paragraph{DAG deduplication.}
The tracer maintains a visited set of turn indices.
When a turn already in $G$ is reached via a second attribution path---the structural pattern of multi-source dependencies---a convergent edge is added to the existing node rather than duplicating it, ensuring $G$ is a DAG.

\paragraph{Path extraction.}
After DAG construction, we apply \emph{relative-threshold extraction} to prune $G$ before evaluation.
At each node $v$, let $s^* = \max_{u}\, s(v, u)$ be the maximum parent edge score.
Given $\alpha \in [0, 1]$, edge $(v, u)$ is retained iff
\begin{equation}
  s(v, u) \;\geq\; \alpha \cdot s^*.
  \label{eq:rel-threshold}
\end{equation}
Unlike an absolute score threshold, this adapts to the local score distribution at each node: attribution scores are small in absolute terms (typically $0.01$--$0.15$ for attention backends) and are not globally calibrated, varying across parts of the graph and across base attributors.
$\alpha$ is therefore a ratio to the strongest parent edge at a node.
Setting $\alpha\!=\!1.0$ reduces to a greedy single chain; lower values admit additional branches when a lower-ranked parent receives comparable attribution to the top candidate.
We set $\alpha\!=\!0.85$; performance is insensitive to this choice within a flat plateau (Appendix~\ref{sec:appendix-alpha}).

\paragraph{Generator-agnostic provenance.}
We instantiate MTCA over a completed transcript, finding which prior turns contain the information needed to account for the target span.
\textsc{Tokengeist} therefore needs only the recorded conversation---not the generator's weights---and the attributing model may either be the generator itself or a different one.
Both settings are useful.
Under a \emph{matched generator}, the attributing model is the one that produced the assistant turns, so the attention \textsc{Tokengeist} reads \emph{is} the attention that produced the text, and an agent tracing its own output can record that attention as the turn is generated rather than recomputing it.
Under a \emph{mismatched generator}, an open-weight reader attributes a transcript produced by a different model, which is what makes \textsc{Tokengeist} applicable to agents served behind closed APIs whose internals are inaccessible---every TauBench result here attributes a \texttt{gpt-4.1} rollout this way.
This surrogate reading is justified empirically: the method ranking and the recursive-over-flat gain are the same in both settings (Section~\ref{sec:matched-generator}).

\subsection{Attribution Backends}
\label{sec:backends}

\textsc{Tokengeist} is agnostic to the choice of $\mathcal{A}$; we evaluate five instantiations.
\textbf{Avg.\ Attention} averages attention weights uniformly across all layers and heads.
\textbf{AT2}~\citep{cohen2025learning} replaces uniform averaging with a pretrained linear probe over attention features.
\textbf{AttnTrace}~\citep{wang2025attntrace} estimates attribution scores via bootstrap subsampling of forward passes.
\textbf{Semantic Retrieval} uses embedding cosine similarity (Appendix~\ref{sec:appendix-semantic}). 
\textbf{ContextCite}~\citep{cohen2024contextcite} scores a source by the target's logit-probability drop under ablation (Appendix~\ref{sec:appendix-contextcite}).
All backends share the same interface and slot into \textsc{Tokengeist} without modifying the recursive tracer.
Each attention-based backend is evaluated under both matched and mismatched generators (Section~\ref{sec:matched-generator}); implementation details are in Appendix~\ref{sec:appendix-impl}.

\section{MTCABench}
\label{sec:mtcabench}

We introduce \textsc{MTCABench}, a benchmark for evaluating multi-turn context attribution.
Each instance contains a multi-turn conversation, a target span in an assistant response, and a gold provenance graph identifying the earlier turns it draws from, testing whether a method recovers both direct evidence and layered dependencies through intermediate assistant turns, tool calls, and tool results.

\subsection{Dataset Details}

\textsc{MTCABench} draws conversations from two agentic function-calling benchmarks, ConFETTI~\citep{alkhouli2025confetti} and TauBench~\citep{yao2024tau} (Table~\ref{tab:dataset_overview}).

\begin{table}[t]
\centering
\small
\begin{tabular}{lrr}
\toprule
 & \textbf{ConFETTI} & \textbf{TauBench} \\
\midrule
Conversations / simulations & 109 & 576 \\
Annotation targets           & 683 & 3{,}287 \\
Avg.\ targets per conversation       & 6.3 & 5.7 \\
Avg.\ turns per conversation         & 16.8 & 27.3 \\
Avg.\ provenance depth       & 3.6 & 6.9 \\
Max.\ provenance depth       & 12 & 17 \\
Domain families              & 10 & 2 \\
\bottomrule
\end{tabular}
\caption{\textbf{\textsc{MTCABench} overview.}
Counts are over every released target. Provenance depth is the longest
target-to-source path in the gold DAG.}
\label{tab:dataset_overview}
\end{table}

\begin{figure}[t]
  \includegraphics[width=\linewidth]{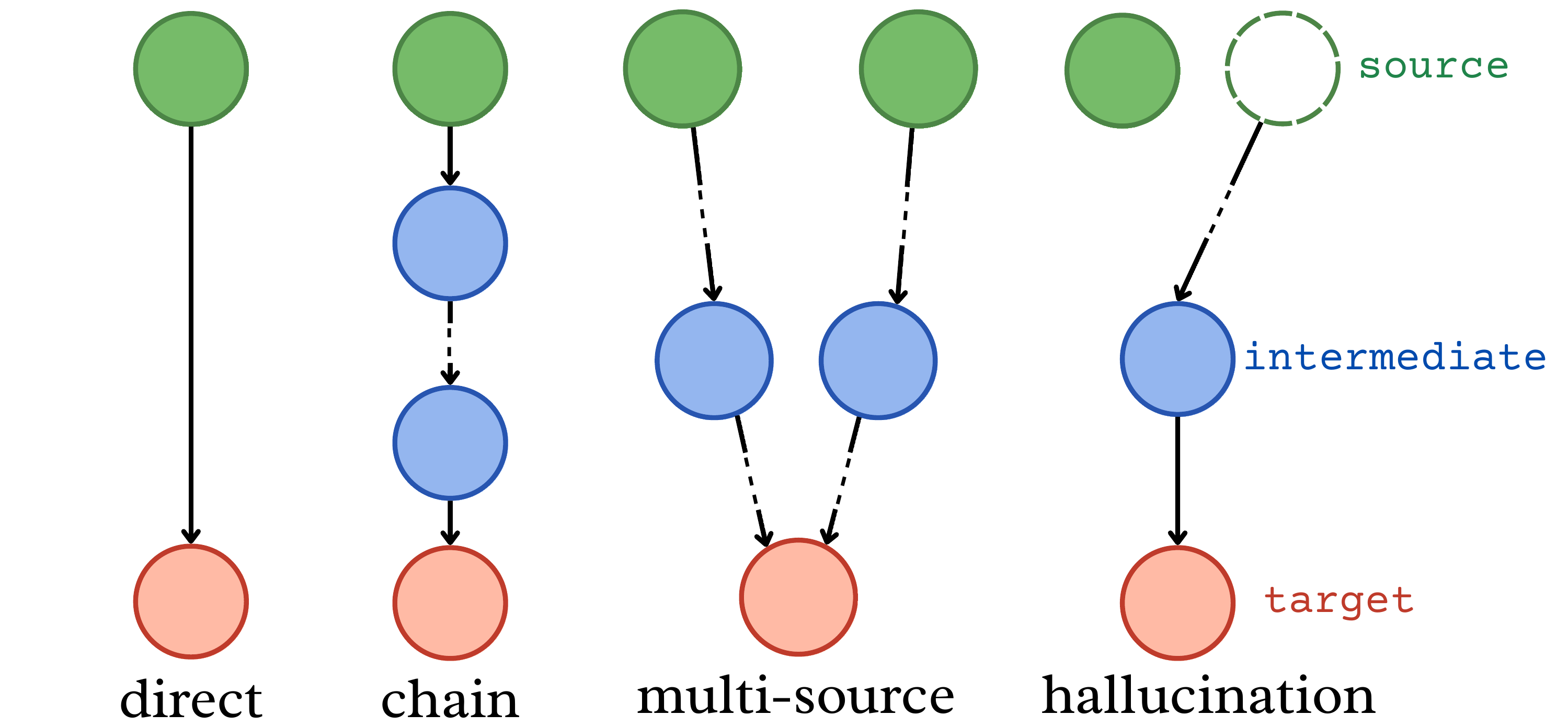}
  \caption{\textbf{Dependency types in \textsc{MTCABench}.} Dashed edges omit intermediate hops for clarity.}
  \label{fig:dependency_types}
\end{figure}

\paragraph{Domain coverage.}
ConFETTI conversations span 10 task families identified by semantic clustering of API call signatures: Accommodation \& Car Rental, Vacation Planning, Business Travel \& HR, Medical Services, Insurance \& Benefits, Lifestyle \& Telecom \& Social, HR Operations, Banking, and Productivity (Table~\ref{tab:domains}).
TauBench adds airline and retail, airline overlapping with ConFETTI's travel clusters, for 11 distinct task families in all.

\paragraph{Dependency types.}
Illustrated in Figure~\ref{fig:dependency_types}, each target carries a base hop-depth tag---\textit{direct} (single hop from source to target) or \textit{chained} (multi-hop through intermediate turns)---and may additionally be tagged \textit{multi-source} (two or more source turns), 
or \textit{hallucination}.
The \textit{hallucination} tag applies to generated spans that are unsupported or factually wrong in context, including wrong-value errors where the model formatted or selected an incorrect value from otherwise relevant evidence.
Per-tag counts and distributions are in Appendix~\ref{sec:appendix-dataset}.

\subsection{Annotation Protocol}
\label{sec:annotation_protocol}
Annotation proceeds in three phases: target identification (phase~1), provenance annotation (phase~2), then validation and selection (phase~3). Phase~1 runs once and fixes the target spans; phase~2 is run independently by four annotators over those targets; phase~3 validates the four annotations and selects the released gold from among them. Full prompts are in Appendix~\ref{sec:appendix-prompts}.

\paragraph{Phase 1: target identification.}
We select 5--8 target spans per conversation from assistant turns, excluding user and tool-result messages since attribution targets must be model-generated.
We use \texttt{GPT-5.4} to do this selection, prioritizing spans that are specific and factual---an ID, date, amount, or paraphrase of a tool result---and traceable to a prior turn through well-defined provenance.
Targets include a mix of dependency types and provenance depths, including hallucination cases. 

\paragraph{Phase 2: provenance annotation.}
Each target span is annotated with a provenance graph tracing all dependencies back to exogenous source nodes, without a depth limit.
Four models annotate the same fixed targets independently: \texttt{GPT-5.4}, then \texttt{GPT-5.6-Sol}, \texttt{GPT-6-Luna} and \texttt{GPT-6.1-Sol} under a refined prompt from step-3b's human review.

\paragraph{Phase 3: validation and selection.}
Because the gold provenance DAGs are themselves LLM-produced, phase~3 validates
them and selects the final gold DAGs in five steps:
(3a) a well-formedness sweep on \texttt{GPT-5.4}'s annotation;
(3b) independent review by three authors of that annotation's graphs for 100 stratified targets;
(3c) the same sweep, plus pairwise agreement, across all four phase-2 model annotations;
(3d) blind human adjudication of 1.1\% of targets on which no pair of annotations reaches agreement of node $F_1 \geq 0.8$;
and (3e) selection of the final gold DAG per target by majority vote over the three re-annotations
and a final sweep on the result.
Full methodology and per-step results are in Appendix~\ref{sec:appendix-annotation-quality}.
\section{Evaluation}
\label{sec:experiments}

Our experiments evaluate the central hypothesis of the paper: flat attribution can identify locally relevant context, but fails to recover layered provenance paths in multi-turn conversations. 
We therefore compare flat attribution methods against recursive variants produced by \textsc{Tokengeist}, using the same underlying attribution scorer.

\subsection{Experimental setup}

We evaluate on four open-weight instruction-tuned models: \texttt{Qwen2.5-7B-Instruct} and \texttt{Llama-3.1-8B-Instruct} (Table~\ref{tab:main_results}); \texttt{DeepSeek-R1-Distill-Qwen-7B} and \texttt{Phi-4-Mini-Instruct} (Appendix~\ref{sec:appendix-extra-models}). GPT-family models are used only for dataset construction, not attribution.
\textsc{Tokengeist} is generator-agnostic: whether assistant turns come from the attributing model itself (Section~\ref{sec:matched-generator}) or from a different model, our findings and method ranking hold. 
For each of the three base scorers introduced in Section~\ref{sec:backends} (\avgattn{}, AT2, AttnTrace), we compare its flat application against its \textsc{Tokengeist} counterpart, isolating the contribution of recursion.
Appendix~\ref{sec:appendix-ablations} reports ablations over recursion depth
$d_{\max}$, branching factor $k$, pruning threshold $\alpha$, and source
granularity, supporting the defaults ($d_{\max}{=}8$, $k{=}3$, $\alpha{=}0.85$,
sentence-level segmentation) used throughout.

\begin{table*}[t]
\centering
\small
\setlength{\tabcolsep}{5pt}
\begin{tabular}{lcccc|cccc}
\toprule
& \multicolumn{4}{c|}{\textit{ConFETTI} ($n$=493)} & \multicolumn{4}{c}{\textit{TauBench} ($n$=3{,}099)} \\
\cmidrule(lr){2-5}\cmidrule(lr){6-9}
Method & Edge F1 & Node F1 & Src.\ Rec. & Span F1 & Edge F1 & Node F1 & Src.\ Rec. & Span F1 \\
\midrule
\multicolumn{9}{l}{\textit{Llama-3.1-8B-Instruct}} \\
\midrule
Flat Attention           &           39.5 &           44.3 &           34.6 &           12.5 &           20.3 &           24.4 &           18.6 &            5.8 \\
Flat AT2                 &           39.6 &           44.8 &           38.1 &           13.3 &           17.0 &           22.1 &           21.0 &            5.8 \\
Flat AttnTrace           &           27.2 &           48.5 &           51.9 &           17.2 &           14.7 &           22.3 &           35.5 &            7.1 \\
\midrule
\textsc{Tg} + Attention  &  \textbf{72.1} &  \textbf{81.4} &           92.6 &  \textbf{47.0} &  \textbf{52.5} &  \textbf{67.8} &  \textbf{89.1} &  \textbf{34.0} \\
\textsc{Tg} + AT2        &           71.2 &           81.1 &  \textbf{92.7} &           47.0 &           40.3 &           60.6 &           88.0 &           31.3 \\
\textsc{Tg} + AttnTrace  &           46.6 &           67.7 &           91.8 &           35.6 &           20.2 &           31.2 &           76.8 &           13.3 \\
\midrule
\multicolumn{9}{l}{\textit{Qwen2.5-7B-Instruct}} \\
\midrule
Flat Attention           &           39.5 &           43.9 &           34.6 &           12.6 &           20.6 &           23.8 &           16.0 &            5.2 \\
Flat AT2                 &           40.7 &           45.1 &           44.8 &           13.5 &           19.1 &           23.6 &           23.3 &            6.0 \\
Flat AttnTrace           &           27.0 &           47.6 &           55.2 &           17.1 &           14.8 &           22.4 &           35.5 &            7.0 \\
\midrule
\textsc{Tg} + Attention  &           73.0 &  \textbf{82.7} &  \textbf{94.1} &  \textbf{48.2} &           55.0 &           70.6 &           88.5 &           35.8 \\
\textsc{Tg} + AT2        &  \textbf{73.4} &           82.6 &           93.0 &           47.3 &  \textbf{62.8} &  \textbf{72.1} &  \textbf{89.3} &  \textbf{40.1} \\
\textsc{Tg} + AttnTrace  &           47.2 &           66.7 &           91.7 &           35.3 &           19.6 &           29.5 &           71.4 &           11.8 \\
\midrule
\multicolumn{9}{l}{\textit{Model-Independent}} \\
\midrule
Flat Semantic            &  26.7 &  39.0 &  58.5 &  10.6 &  10.3 &  18.9 &  39.8 &   5.7 \\
\textsc{Tg} + Semantic   &  37.4 &  53.9 &  80.7 &  30.0 &  14.7 &  26.3 &  76.5 &  10.7 \\
\midrule
\multicolumn{9}{l}{\textit{Matched generator}$^{\dagger}$} \\
\midrule
Flat Attention \textit{(Llama)}          &  52.5 &  57.2 &  70.1 &  14.5 &  32.1 &  38.5 &  31.3 &   5.1 \\
\textsc{Tg} + Attention \textit{(Llama)} &  83.0 &  88.4 &  97.5 &  47.5 &  58.7 &  69.9 &  89.3 &  20.4 \\
Flat Attention \textit{(Qwen)}           &  49.6 &  53.3 &  63.9 &  12.9 &  34.7 &  37.6 &  27.0 &   4.1 \\
\textsc{Tg} + Attention \textit{(Qwen)}  &  79.8 &  86.0 &  93.7 &  46.5 &  65.0 &  72.2 &  90.8 &  15.7 \\
\bottomrule
\end{tabular}
\caption{\textbf{Flat versus recursive attribution on \textsc{MTCABench}.}
Percentages, mean of 3 runs; \textsc{Tg} = \textsc{Tokengeist}, Attention = \avgattn{}; bold marks the best recursive method per metric per model.
AT2 and \avgattn{} are deterministic; AttnTrace's standard deviation (${\lesssim}1.1$ points) never changes rankings (Appendix~\ref{sec:appendix-experiments}).
Semantic reads no model internals and is reported once, as the mean over the four models (Appendix~\ref{sec:appendix-semantic}); Phi-4-Mini-Instruct and DeepSeek-R1-Distill-Qwen-7B follow the same pattern (Table~\ref{tab:appendix_main_results}).
$^{\dagger}$Each model attributes transcripts it generated itself (Section~\ref{sec:matched-generator}); regenerating one succeeds only where its assistant turns stay close enough to the record for the pre-loaded user turns to still follow, and we score only targets whose whole gold provenance path survives---
$n$=291 / 242 (ConFETTI), 398 / 319 (TauBench), (Llama / Qwen).
AT2 and AttnTrace behave the same way (Appendix~\ref{sec:appendix-selfgen}).}
\label{tab:main_results}
\end{table*}

\subsection{Metrics}
\label{sec:metrics}

We evaluate each predicted provenance graph $\hat{G}$ against a gold provenance DAG $G^*$ on four metrics.
Because MTCA represents dependencies as directed edges between conversation events, our primary structural metric is \textbf{Edge F1}: precision, recall and F1 over directed dependency edges, where $(u,v)$ is correct if both $\hat{G}$ and $G^*$ include $v$ as a provenance source of $u$.
\textbf{Node F1} scores whether a method found the correct supporting turns regardless of how it connected them, which separates evidence-retrieval failures from graph-structure failures.
\textbf{Source recall} is recall over gold leaf sources---exogenous nodes with no further dependencies---and measures whether a method reaches the origin of the information rather than stopping at an intermediate turn.
\textbf{Span F1} is word-overlap F1 between the attributed spans stored in predicted nodes and the gold span annotations, macro-averaged over all gold-annotated turns, with turns absent from the predicted graph contributing F1~$=0$.
It is strictly harder than Node F1, since both the supporting turn (\emph{coverage}) and the sub-turn excerpt (\emph{granularity}) must be correct.

\begin{figure*}[tb]
  \centering
  \includegraphics[width=\linewidth]{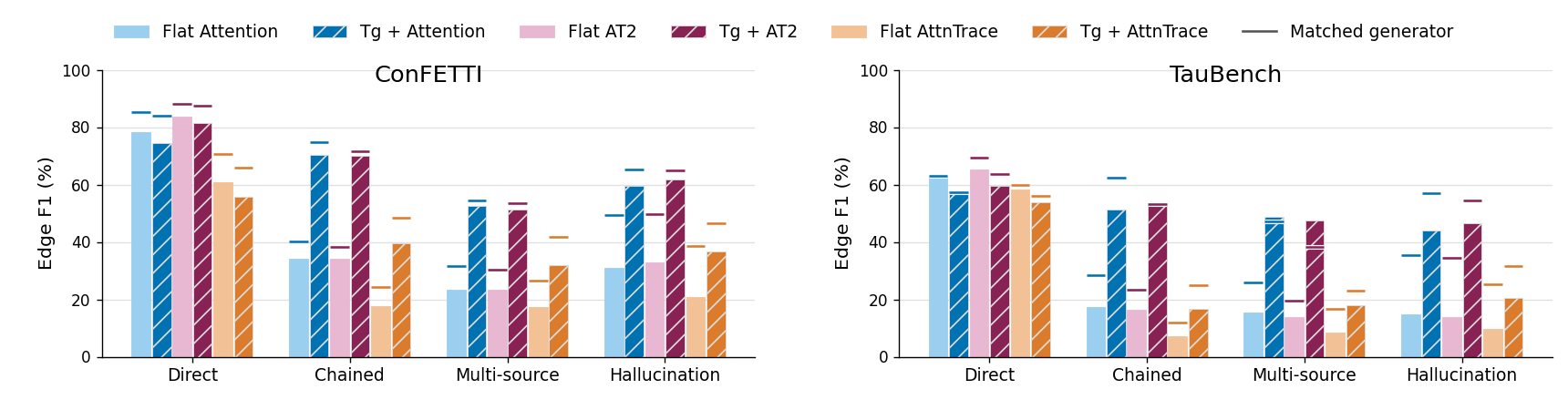}
  \caption{\textbf{Edge F1 by dependency type} on ConFETTI (left) and TauBench (right).
  Light bars: flat methods; dark bars: \textsc{Tg} (= \textsc{Tokengeist}).
  Ticks mark the same method under a matched generator (Section~\ref{sec:matched-generator}), which admits a smaller target set, so ticks are indicative rather than an equivalent comparison.}
  \label{fig:dep_type_results}
\end{figure*}

\subsection{Tokengeist recovers collapsed provenance}

Recursive attribution outperforms its flat counterpart on every metric, on both datasets, and for all four models (Table~\ref{tab:main_results}).\footnote{Each split reserves a held-out slice of roughly 190 targets (30 ConFETTI conversations, 33 TauBench simulations), leaving the 493 and 3{,}099 targets of Table~\ref{tab:main_results}. The pruning threshold $\alpha$ and the branching factor $k$ were selected on the ConFETTI slice alone; the TauBench slice is used only for the $d_{\max}$ sweep (Appendix~\ref{sec:appendix-ablations}). No reported score is computed on data used for selection; sampling details are in Appendix~\ref{sec:appendix-impl}.}
With the \avgattn{} and AT2 backends, recursion improves Edge F1 by 23--44 percentage points.
Even AttnTrace, the weakest flat baseline, gains a median of 16 Edge F1 points across models and datasets.
\textsc{Tokengeist} is therefore a drop-in upgrade for any base scorer.

Source recall on TauBench rises from 18\% to 87\% on average across models for \avgattn{} and AT2.
Flat attribution stops at the wrong turn prematurely, and \textsc{Tokengeist} overcomes \textit{provenance collapse} by tracing each intermediate assistant turn back through its own context, so one long-range attribution query becomes a chain of short-range ones.
The result is provenance that reaches the originating user or tool event, not just the nearest assistant turn.

\paragraph{Recursion drives the gain.} Three controls isolate where the gain comes from. First, the deterministic structural edge alone (Flat+Struct, Appendix~\ref{sec:appendix-struct-ablation}) accounts for roughly a third to half of the Edge F1 gain, but for almost none of the source-recall gain: on TauBench it adds at most 4.0 points where recursion adds 32.7--70.7. Second, the Semantic Retrieval baseline (Table~\ref{tab:main_results}) shows recursive traversal lifts source recall for a scorer that reads no model internals, from 58.5 to 80.7 on ConFETTI and from 39.8 to 76.5 on TauBench. However, it only reaches 37.4 on Edge F1. Third, the counterfactual necessity baseline (ContextCite, Appendix~\ref{sec:appendix-contextcite}) recovers 65.1--67.0 of the 65.7--73.0 Edge F1 that \textsc{Tg}~+~Attention attains across the four attribution models. Recovering the originating event is therefore largely attributable to the recursion, and intermediate turns to the quality of the base attributor.
The raw signal shows the same split: the share of gold source turns receiving any non-zero score falls below 20\% two hops from the target for flat attribution, but holds above 50\% at every depth for \textsc{Tokengeist} (Appendix~\ref{sec:appendix-signal-depth}).


\subsection{Performance by dependency type}

Next, we stratify performance by dependency type, testing whether gains from recursion are concentrated in the cases that motivate MTCA: chained and multi-source dependencies. Figure~\ref{fig:dep_type_results} shows
on direct (single-hop) dependencies, flat methods perform well since no intermediate turn needs to be traversed.
On chained dependencies, flat methods drop to under 35\% Edge F1 while \textsc{Tokengeist} reaches 70\%.
Multi-source targets are the hardest category: flat attention averages 23\% Edge F1 while \textsc{Tokengeist} reaches 52\%.

Hallucination targets illustrate the diagnostic role of recursive tracing: for an incorrectly generated span, the goal is to find the context that primed or propagated it, and recursive attribution substantially improves localization of where an erroneous value entered the trace for a reviewer or downstream verifier (Appendix~\ref{sec:appendix-hallucination}).

\subsection{Matched and mismatched generators}
\label{sec:matched-generator}

The ranking of methods is the same whether the attribution model matches the generator.
We regenerate every assistant turn of \textsc{MTCABench} with the attributing model (a \emph{replay}) and re-evaluate on targets whose gold provenance path the replay reproduces in full.
Recursive attribution still beats flat attribution in all four replays and for every backend (Table~\ref{tab:main_results}), with the gain concentrated on chained and multi-source dependencies.
Appendix~\ref{sec:appendix-selfgen} reports the full results.
\section{Conclusion}

We introduced \emph{multi-turn context attribution}, which recasts
provenance recovery as a recursive traversal over a DAG of conversation events,
and instantiated it as \textsc{Tokengeist}, an attribution-method-agnostic
framework we will release alongside \textsc{MTCABench}. Across four
open-weight models, \textsc{Tokengeist} raises source recall from 11--38\% to
82--94\%. It
roughly doubles edge~F1 over flat baselines, exposing \emph{provenance collapse}
as a systematic failure of single-pass attribution.

The recursive structure invites tighter coupling with the agent loop:
provenance graphs could be consumed online by planners and self-correction
mechanisms, letting agents detect when a claim rests on a shaky intermediate
turn and repair it before acting. Scaling MTCA to longer
horizons---persistent memory, cross-session histories, and multi-agent
transcripts---will require attribution backends that amortize cost across
overlapping subgraphs and prune more aggressively than our depth-bounded BFS.

\clearpage

\section*{Limitations}
\label{sec:limitations}

We outline the main limitations of \textsc{Tokengeist} and \textsc{MTCABench}.

\paragraph{Coverage of \textsc{MTCABench}.}
\textsc{MTCABench} is built from two English-language sources---ConFETTI~\citep{alkhouli2025confetti} and TauBench~\citep{yao2024tau} (airline and retail domains)---and is text-only.
It does not cover multilingual dialogue, multimodal inputs, very long persistent-memory sessions, or multi-agent transcripts.
Generalization of our results to other domains, languages, and modalities is plausible but unverified.

In agentic traces deep provenance is almost always routed through tool calls, so depth and tool structure are confounded in \textsc{MTCABench}: phase~1 prioritizes target spans that carry an ID or argument or paraphrase a tool result (Section~\ref{sec:annotation_protocol}).
Our claims are therefore scoped to deep provenance in agentic conversations---the deterministic tool-result edge is not what produces the gain (Appendix~\ref{sec:appendix-struct-ablation}), but we leave investigating how recursive tracing transfers to ordinary multi-turn dialogue carrying no tool scaffold to future work.

\paragraph{Quality is bounded by the base attribution method.}
\textsc{Tokengeist} composes multiple calls to a single-turn attribution scorer $\mathcal{A}$ and inherits the strengths and weaknesses of that scorer.
This is most visible for AttnTrace, which performs poorly even in its flat form (e.g., 8.7--27.2 edge~F1 on \textsc{MTCABench}); \textsc{Tg}+AttnTrace correspondingly underperforms \textsc{Tg}+Attention and \textsc{Tg}+AT2 by a wide margin, especially on TauBench where its flat attribution scores are weakest.
That said, our recursive solution still improves over the flat baseline in every model$\times$backend cell we report (Table~\ref{tab:main_results}), so the benefit of recursive tracing is robust to the choice of backend even when the absolute numbers are not.
Better single-turn attribution methods, when they appear, should compose with \textsc{Tokengeist} without modification and lift the ceiling further.

\paragraph{Reproducibility of attribution backends.}
AT2 and \avgattn{} are deterministic up to float16 GPU
non-determinism ($\lesssim 0.2$~point variance, never affecting method rankings),
while AttnTrace is stochastic by design and we follow the upstream
implementation in not fixing the bootstrap seed; fixing the seed would not
eliminate the GPU non-determinism shared with the other backends.

\paragraph{Model scope.}
Since our backends require access to model internals (attention weights or hidden states), we evaluate on four open-weight 7--8B instruction-tuned models.
This constrains the \emph{attribution} model only: the model that generated the conversation may be closed, as it is throughout our TauBench split.
The specific models are further constrained by AT2 probe availability: AT2 requires a pretrained linear probe per model architecture, and we use the four architectures for which probes are publicly released~\citep{cohen2025learning}.
We do not report results on much larger open-weight models, closed-weight API models, or base (non-instruction-tuned) checkpoints, and we cannot rule out that provenance collapse manifests differently at other scales or after different post-training procedures.


\paragraph{Turn-level granularity.}
DAG nodes in \textsc{Tokengeist} are conversation turns rather than spans.
This choice is motivated in Section~\ref{sec:tokengeist}, but it caps within-turn localization: a method that selects the correct turn but returns its full content is penalized by Span~F1, which is consistently the lowest of our four metrics.
Span- or sentence-level recursion is a natural extension but requires resolving what it means to recurse on a sub-turn fragment that was generated as part of a single autoregressive pass.
Prior work on citation granularity points in both directions: subsentence citations can improve verifiability by localizing exactly which generated claims are supported~\citep{cao-wang-2024-verifiable}, while finer citation units are not uniformly better and may reduce attribution quality when they break useful semantic context~\citep{wang2026finer}.
For multi-turn provenance graphs, finer nodes would also make the graph messier: longer conversations would yield many more nodes and edges, edge weights would vary across a wider range of local contexts, and normalizing or thresholding those weights across heterogeneous sub-turn fragments would be harder.
Developing span-level provenance graphs that preserve interpretability without sacrificing graph tractability or score calibration is therefore future work.

\paragraph{Hyperparameter sensitivity.}
The defaults reported in the main text ($k\!=\!3$, $d_{\max}\!=\!8$, $\alpha\!=\!0.85$, $\theta\!=\!0$) were selected on held-out slices disjoint from every reported score---$\alpha$ and $k$ on ConFETTI alone, $d_{\max}$ on both---and reused across all models and datasets.
We report ablations over each in Section~\ref{sec:appendix-ablations}, but performance under distribution shift to new conversation styles may benefit from re-tuning.

\paragraph{Computational cost.}
Recursive attribution costs $O(k^{d_{\max}})$ calls to $\mathcal{A}$ in the worst case.
For attention-aggregation backends each call is a single forward pass, but for AttnTrace, which adds bootstrap subsampling, the cost compounds.
When the agent attributes its own output, the first call is free---its attention can be recorded as the turn is generated (Section~\ref{sec:matched-generator})---but every recursive step explains a different turn and still costs a forward pass.
Practitioners running \textsc{Tokengeist} on very long agentic transcripts or in latency-sensitive settings will need to tune $k$, $d_{\max}$, and the pruning threshold $\alpha$ accordingly.

\paragraph{Ground-truth provenance is a model-produced construct.}
Gold provenance DAGs are model-produced; no human authored a graph from scratch.
Intrinsic to the task, there is no oracle provenance for a conversation: our metrics
measure agreement with a carefully constructed annotation standard, not with the model's
internal causal graph. Appendix~\ref{sec:appendix-annotation-quality} reports the
validation in full.
\section*{Impact Statement}
This paper is about understanding where a model's answer comes from in a multi-turn conversation.
In many real deployments, a response is not produced from a single prompt alone: it may depend on user instructions, earlier assistant summaries, or tool outputs spread across many turns.
When that information is copied, condensed, or transformed along the way, it becomes difficult to tell what the final answer is actually grounded in.
Our work aims to make this flow of information easier to inspect.

The main benefit of this work is transparency and oversight.
A method for multi-turn attribution can help developers and researchers see whether a particular claim is grounded in an original source, an intermediate assistant message, a tool result, or a spurious piece of context.
This can be useful for debugging long-context systems, analyzing agent traces, building better benchmarks, and studying when attribution methods stop too early at a nearby summary instead of tracing back to the true source.
More broadly, we hope this encourages systems to preserve better provenance information across conversations and tool-use workflows.

At the same time, attribution is not a substitute for correctness or safety.
A recovered provenance path can help explain what context appears to have influenced a response, but it does not prove that the response is true, complete, or safe to act on.
There is also a risk that users may over-trust attribution outputs, especially when the underlying attribution scores are noisy or incomplete.
Because multi-turn traces may include private user messages, retrieved documents, or tool outputs, any practical use of this kind of method should include appropriate privacy, access-control, and data-retention safeguards.

Finally, MTCABench reflects the domains and annotation choices used to build it.
It should be treated as a useful starting point for studying multi-turn provenance, not as a complete map of all possible conversational settings.
Overall, we see this work as a step toward more inspectable model interactions: it helps show how information moves through a conversation, but it should be used alongside careful evaluation, privacy review, and domain-specific safety testing.
\bibliography{custom}
\appendix
\setcounter{topnumber}{4}
\setcounter{totalnumber}{6}
\setcounter{dbltopnumber}{4}
\renewcommand{\topfraction}{0.9}
\renewcommand{\bottomfraction}{0.5}
\renewcommand{\dbltopfraction}{0.9}
\renewcommand{\textfraction}{0.07}
\renewcommand{\floatpagefraction}{0.6}
\renewcommand{\dblfloatpagefraction}{0.6}
\section{Experiment Details}
\label{sec:appendix-experiments}

\subsection{Implementation Details}
\label{sec:appendix-impl}

\paragraph{Software stack.}
\textsc{Tokengeist} is implemented in Python~3.10 on top of HuggingFace \texttt{transformers}~4.51.3, \texttt{accelerate}~1.1.1, and PyTorch~2.10 (CUDA~13.0 runtime, cuDNN~9.2).
Sentence segmentation uses NLTK \texttt{punkt\_tab}.
All experiments use the same code path for flat and recursive attribution: a flat run is exactly the special case $d_{\max}{=}1$ of the recursive tracer, so the only difference between paired (flat, recursive) numbers in Table~\ref{tab:main_results} is the recursion depth.

\paragraph{Models.}
We evaluate on four open-weight instruction-tuned causal language models, loaded in \texttt{float16} on a single GPU per process:
\texttt{Qwen/Qwen2.5-7B-Instruct},
\texttt{meta-llama/Llama-3.1-8B-Instruct},
\texttt{microsoft/phi-4-mini-instruct},
and
\texttt{deepseek-ai/\allowbreak DeepSeek-R1-Distill-Qwen-7B}.
All models are used as released, with no fine-tuning.
For all backends we set \texttt{output\_hidden\_states=True} on the attribution forward pass so that the same hidden states can be reused across the attention extraction step and downstream attribution scoring.

For ConFETTI, conversation history is presented to the model in the flat \texttt{Role: content} format used by the underlying tracer; for TauBench, where the upstream agent was driven through the model's native chat template, we set \texttt{use\_chat\_template=True} so that the attribution forward pass sees inputs in the same format as the original generation, including tool-call delimiters.
Tool-call names and JSON arguments are serialized into the textual content of the issuing assistant turn so that attention from later turns can resolve to function names and argument values.

\paragraph{Attribution backends.}
We use five single-turn attribution backends, all of which expose the same interface and can be slotted into \textsc{Tokengeist} without changes to the recursive tracer:
\begin{itemize}
  \item \textbf{Avg.\ Attention.} Native attention scores averaged uniformly across all transformer layers and attention heads, then summed over each candidate source span and averaged over the target tokens (Appendix~\ref{sec:appendix-attn}).
  \item \textbf{AT2}~\citep{cohen2025learning}. A pretrained linear probe over attention features, loaded from the public \texttt{madrylab/at2-*} HuggingFace probes corresponding to each evaluated model.
  No additional training is performed.
  \item \textbf{AttnTrace}~\citep{wang2025attntrace}. Bootstrap-style attention attribution with $B{=}30$ subsampled forward passes, subsampling fraction $q{=}0.4$, and top-$k_{\text{attn}}{=}5$ token aggregation per segment. These are the original AttnTrace defaults.
  \item \textbf{Semantic Retrieval.} Embedding cosine similarity, reading no model internals; settings in Appendix~\ref{sec:appendix-semantic}.
  \item \textbf{ContextCite}~\citep{cohen2024contextcite}. A sparse linear surrogate over ablated contexts; settings in Appendix~\ref{sec:appendix-contextcite}.
\end{itemize}
The first three are the attention-based backends reported in Tables~\ref{tab:main_results} and~\ref{tab:appendix_main_results}; the last two are controls on the recursion itself and are reported in Table~\ref{tab:semantic-ablation}, as the mean over the four attribution models, with ContextCite broken out per model in Table~\ref{tab:contextcite-permodel} (Appendix~\ref{sec:appendix-backends}).
In every case the same backend is used for the corresponding flat baseline and \textsc{Tg}+backend rows, isolating the effect of recursion.

\paragraph{\textsc{Tokengeist} hyperparameters.}
Unless stated otherwise, we use the following configuration for every method and benchmark:
maximum recursion depth $d_{\max}{=}8$ (flat: $d_{\max}{=}1$);
per-step branching factor $k{=}3$;
score threshold $\theta{=}0$ (no pre-filtering of low-score candidates before top-$k$);
source segmentation granularity \texttt{sentence};
DAG node granularity \texttt{turn} (per-turn aggregation by max over span scores);
tool-result $\to$ tool-invocation structural edges with fixed score $1.0$;
and relative-threshold path extraction with $\alpha{=}0.85$ applied uniformly to all predicted graphs before scoring (Section~\ref{sec:appendix-alpha}).
\textbf{Span F1} (Section~\ref{sec:metrics}) is computed on the same DAGs without re-tracing: per-turn node text is tokenized by whitespace and matched against the gold span annotation.

\paragraph{Hyperparameter holdout.}
All hyperparameters above ($k$, $d_{\max}$, $\alpha$, $\theta$) are selected on a holdout split that is disjoint from the data used for every reported score.
The holdout is drawn by sampling whole conversations (ConFETTI) and whole simulations (TauBench) uniformly at random with Python's \texttt{random.Random} under a fixed seed of~42, so that no target from a held-out conversation can leak into the evaluation split.
This yields 30 of 109 ConFETTI conversations (190 targets) and 33 of 576 TauBench simulations (187 targets).
The TauBench slice is drawn by sampling 26 \texttt{sim\_id}s; because 127 of the 449 distinct \texttt{sim\_id}s occur in both the airline and retail domains and denote different conversations, sampling an id holds out every conversation carrying it, which is why 26 ids yield 33 simulations.
Table~\ref{tab:main_results} and all other reported results are computed on the complement: 493 ConFETTI targets (79 conversations) and 3{,}099 TauBench targets (543 simulations).

\paragraph{Annotation pipeline.}
Ground-truth provenance DAGs are produced by the pipeline described in Section~\ref{sec:annotation_protocol}.
Phase~1 (target identification) uses \texttt{GPT-5.4} and runs once; the targets it selects are held fixed thereafter.
Phase~2 (provenance annotation) is run four times independently---by \texttt{GPT-5.4} (escalating to \texttt{GPT-5.4-Pro} on repeated failure) and by \texttt{GPT-5.6-Sol}, \texttt{GPT-6-Luna} (both at the deployment's default reasoning effort) and \texttt{GPT-6.1-Sol} (high)---and the released gold is assembled from those four by majority vote, with the residual on which no two agree adjudicated by hand (Appendix~\ref{sec:appendix-annotation-quality}).
\texttt{GPT-5.4} ran first, under an earlier version of the phase-2 prompt.
Generation uses each deployment's default sampling parameters with JSON-mode output, retried up to three times on transient errors; a target that still fails in the initial pass is escalated to \texttt{GPT-5.4-Pro}.
The labeling pipeline is fully separate from the attribution pipeline: GPT-family models are never used as attribution models or scorers.

\paragraph{Reproducibility.}
All results in the paper are averaged over 3 independent runs.
The two AT2 backends (flat and recursive) are deterministic given the same model checkpoint and inputs.
\avgattn{} is deterministic up to float16 GPU non-determinism (we observe $\lesssim 0.2$ point variance on individual metrics across re-runs on the same hardware, which never changes method rankings).
AttnTrace is stochastic by design (bootstrap sampling); we use NumPy's default RNG without a fixed seed, which mirrors the original implementation.
Across all 4 models and both flat and recursive AttnTrace variants, measured run-to-run standard deviation on TauBench is $\lesssim 0.6$ points on all four reported metrics.
On ConFETTI, Edge~F1, Node~F1, and Span~F1 are within $0.7$ points; Src.\ Rec.\ standard deviation reaches up to $1.1$ points (largest cases: Qwen2.5-7B flat and Phi-4-Mini recursive), driven by the bootstrap resampling interacting with the sparser source-recall signal on this benchmark.
In no case does run-to-run variance change method rankings.
Code, model checkpoints, label files, and per-target predicted graphs will be released with the camera-ready version.

\subsection{Additional Model Results}
\label{sec:appendix-extra-models}

Table~\ref{tab:appendix_main_results} reports the same flat-versus-recursive comparison as Table~\ref{tab:main_results} on the remaining two evaluated models, \texttt{Phi-4-Mini-Instruct} and \texttt{DeepSeek-R1-Distill-Qwen-7B}.
The same headline patterns observed in the main table also hold here: recursive \textsc{Tokengeist} substantially outperforms every flat baseline, source recall on TauBench crosses the 20\%-to-80--90\% threshold, and AT2 leads \avgattn{} on TauBench on both models.
\textsc{Tg}+AttnTrace remains the weakest recursive variant on both models, consistent with the backend-dependence discussion in the Limitations.

\begin{table*}[t]
\centering
\small
\setlength{\tabcolsep}{5pt}
\begin{tabular}{lcccc|cccc}
\toprule
& \multicolumn{4}{c|}{\textit{ConFETTI} ($n$=493)} & \multicolumn{4}{c}{\textit{TauBench} ($n$=3{,}099)} \\
\cmidrule(lr){2-5}\cmidrule(lr){6-9}
Method & Edge F1 & Node F1 & Src.\ Rec. & Span F1 & Edge F1 & Node F1 & Src.\ Rec. & Span F1 \\
\midrule
\multicolumn{9}{l}{\textit{Llama-3.1-8B-Instruct}} \\
\midrule
Flat Attention           &           39.5 &           44.3 &           34.6 &           12.5 &           20.3 &           24.4 &           18.6 &            5.8 \\
Flat AT2                 &           39.6 &           44.8 &           38.1 &           13.3 &           17.0 &           22.1 &           21.0 &            5.8 \\
Flat AttnTrace           &           27.2 &           48.5 &           51.9 &           17.2 &           14.7 &           22.3 &           35.5 &            7.1 \\
\midrule
\textsc{Tg} + Attention  &  \textbf{72.1} &  \textbf{81.4} &           92.6 &  \textbf{47.0} &  \textbf{52.5} &  \textbf{67.8} &  \textbf{89.1} &  \textbf{34.0} \\
\textsc{Tg} + AT2        &           71.2 &           81.1 &  \textbf{92.7} &           47.0 &           40.3 &           60.6 &           88.0 &           31.3 \\
\textsc{Tg} + AttnTrace  &           46.6 &           67.7 &           91.8 &           35.6 &           20.2 &           31.2 &           76.8 &           13.3 \\
\midrule
\multicolumn{9}{l}{\textit{Qwen2.5-7B-Instruct}} \\
\midrule
Flat Attention           &           39.5 &           43.9 &           34.6 &           12.6 &           20.6 &           23.8 &           16.0 &            5.2 \\
Flat AT2                 &           40.7 &           45.1 &           44.8 &           13.5 &           19.1 &           23.6 &           23.3 &            6.0 \\
Flat AttnTrace           &           27.0 &           47.6 &           55.2 &           17.1 &           14.8 &           22.4 &           35.5 &            7.0 \\
\midrule
\textsc{Tg} + Attention  &           73.0 &  \textbf{82.7} &  \textbf{94.1} &  \textbf{48.2} &           55.0 &           70.6 &           88.5 &           35.8 \\
\textsc{Tg} + AT2        &  \textbf{73.4} &           82.6 &           93.0 &           47.3 &  \textbf{62.8} &  \textbf{72.1} &  \textbf{89.3} &  \textbf{40.1} \\
\textsc{Tg} + AttnTrace  &           47.2 &           66.7 &           91.7 &           35.3 &           19.6 &           29.5 &           71.4 &           11.8 \\
\midrule
\multicolumn{9}{l}{\textit{Phi-4-mini-Instruct}} \\
\midrule
Flat Attention           &           38.9 &           43.9 &           37.5 &           12.3 &           18.7 &           21.6 &           15.6 &            4.9 \\
Flat AT2                 &           40.7 &           46.4 &           37.7 &           13.9 &           18.0 &           22.8 &           19.0 &            5.5 \\
Flat AttnTrace           &           19.1 &           32.3 &           40.7 &            9.5 &            8.7 &           13.3 &           27.7 &            3.9 \\
\midrule
\textsc{Tg} + Attention  &           70.5 &           80.1 &  \textbf{91.6} &           46.5 &           52.3 &           67.8 &  \textbf{87.4} &           36.4 \\
\textsc{Tg} + AT2        &  \textbf{73.4} &  \textbf{83.0} &           91.3 &  \textbf{47.6} &  \textbf{61.5} &  \textbf{71.4} &           80.3 &  \textbf{40.1} \\
\textsc{Tg} + AttnTrace  &           32.2 &           45.9 &           74.9 &           24.3 &            9.2 &           14.6 &           38.3 &            4.6 \\
\midrule
\multicolumn{9}{l}{\textit{DeepSeek-R1-Distill-Qwen-7B}} \\
\midrule
Flat Attention           &           35.6 &           40.2 &           28.0 &           10.6 &           15.6 &           18.5 &           11.4 &            3.6 \\
Flat AT2                 &           39.7 &           45.1 &           35.6 &           13.7 &           18.2 &           22.2 &           16.1 &            5.1 \\
Flat AttnTrace           &           24.5 &           38.2 &           39.4 &           11.3 &            9.2 &           13.8 &           13.6 &            2.9 \\
\midrule
\textsc{Tg} + Attention  &           65.7 &           76.1 &           89.5 &           45.7 &           50.0 &           67.8 &           82.5 &  \textbf{36.1} \\
\textsc{Tg} + AT2        &  \textbf{71.4} &  \textbf{81.0} &  \textbf{93.1} &  \textbf{47.0} &  \textbf{53.4} &  \textbf{69.1} &  \textbf{87.3} &           35.5 \\
\textsc{Tg} + AttnTrace  &           44.0 &           62.6 &           88.0 &           33.6 &           34.2 &           53.1 &           72.7 &           27.7 \\
\bottomrule
\end{tabular}
\caption{\textbf{Flat versus recursive attribution on \textsc{MTCABench} (additional models).}
Same setup, metrics, and bolding convention as Table~\ref{tab:main_results}.
All scores are percentages; mean across 3 independent runs.}
\label{tab:appendix_main_results}
\end{table*}

\subsection{Attention Extraction Details}
\label{sec:appendix-attn}

The \avgattn{} backend reads attention weights directly from the attribution forward pass rather than calling \texttt{model(..., output\_attentions=True)}, which would force the slower \texttt{eager} attention path and require materialising the full $O(L^2)$ matrix.
We instead capture per-layer hidden states via \texttt{output\_hidden\_states=True}, then re-project queries and keys at attribution time using the model's own Q/K projection layers, rotary embeddings (\texttt{apply\_rotary\_pos\_emb}), and GQA key-expansion (\texttt{repeat\_kv}), restricting the attention slice to the target token range $[t_{\text{start}}, t_{\text{end}})$.
We then average uniformly over all $L$ transformer layers and all $H$ attention heads:
\begin{equation}
  \bar{A}_{ij} \;=\; \frac{1}{L \cdot H} \sum_{\ell=1}^{L} \sum_{h=1}^{H} A^{(\ell, h)}_{ij},
  \label{eq:attn-avg}
\end{equation}
where $A^{(\ell, h)} \in \mathbb{R}^{|\bar{y}| \times |X|}$ is the post-softmax attention from target tokens to context tokens at layer $\ell$ and head $h$.
The per-source score is then $s_k = \tfrac{1}{|\bar{y}|}\sum_{i \in \bar{y}} \sum_{j \in \text{src}_k} \bar{A}_{ij}$, with span ranges $\text{src}_k$ taken from the sentence-level segmentation of the context.
Scores are non-negative because individual $A^{(\ell, h)}_{ij}$ are post-softmax.

We support three dispatch paths (\texttt{llama}, \texttt{qwen2}, \texttt{phi3}) through a small dispatch layer that looks up the right \texttt{apply\_rotary\_pos\_emb} / \texttt{repeat\_kv} helpers per architecture; \texttt{DeepSeek-R1-Distill-Qwen-7B} routes to the \texttt{qwen2} path automatically via name matching, since it shares the Qwen2 attention architecture.
We use uniform layer-and-head averaging rather than the popular early-layer-only or last-layer-only heuristics; the AT2 probe is included precisely to compare against a learned reweighting of the same per-head features.

\paragraph{Tool-result handling.}
Tool-result turns contain no model-generated tokens, so their attention weights are not informative.
When recursion reaches a tool-result turn, \textsc{Tokengeist} short-circuits the attribution call: it adds a deterministic structural edge with score $1.0$ to the nearest preceding assistant turn (the issuer of the call), then resumes normal attribution from that assistant turn.
This encodes the causal chain \textit{target} $\to$ \textit{tool result} $\xrightarrow{\textit{struct.}}$ \textit{tool invocation} $\to$ \textit{argument sources} without giving structural edges spurious soft weight.

\subsection{Computation Details}
\label{sec:appendix-compute}

\paragraph{Hardware.}
All experiments run on two Ubuntu~22.04 servers, each equipped with 4$\times$ NVIDIA A100 80GB PCIe GPUs (8 GPUs in total; driver 580.159.03, CUDA toolkit 13.2).
Each attribution process is pinned to a single GPU via \texttt{CUDA\_VISIBLE\_DEVICES} (a 7--8B-parameter model loaded in \texttt{float16} comfortably fits on one card), and conversations are sharded round-robin across the available GPUs by the runner scripts; each process loads its own model copy.
We do not use \texttt{flash-attn}: the attribution forward pass relies on cached hidden states via \texttt{output\_hidden\_states=True}, which is incompatible with the fused flash-attention kernel, so all four models run with the \texttt{eager} attention implementation.

\paragraph{Per-target runtime.}
The asymptotic cost of \textsc{Tokengeist} is $O(k^{d_{\max}})$ attribution calls per target in the worst case, but the effective frontier shrinks rapidly because (a) recursion terminates at exogenous user/system turns, (b) tool-result $\to$ tool-invocation hops are zero-cost (structural edges, no model call), and (c) the visited set ensures each (turn, span) pair is attributed at most once per DAG.
With $k{=}3$ and $d_{\max}{=}8$, observed DAG sizes on ConFETTI are 4--12 nodes (mean ${\approx}7$) and on TauBench 8--30 nodes (mean ${\approx}18$), reflecting the longer tool-call chains in agentic simulations.

Mean wall-clock attribution time per target (single A100): on ConFETTI, flat \avgattn{} and AT2 each take ${\approx}0.2$\,s (probe inference adds negligible overhead), and flat AttnTrace takes ${\approx}3$\,s ($B{=}30$ forward passes dominate; range 2.7--3.5\,s across models).
Phi-4-mini is the fastest model evaluated (0.14\,s flat \avgattn{}) due to its smaller hidden dimension; the three 7--8B models are within 0.06\,s of each other.
Recursive variants take ${\approx}0.6$\,s (\avgattn{}, AT2) and ${\approx}12$\,s (AttnTrace) on ConFETTI, and ${\approx}1.2$\,s, ${\approx}1.4$\,s, and ${\approx}17$\,s respectively on TauBench (longer tool-call chains increase the effective DAG size).
This corresponds to a 3--4$\times$ overhead over the flat baseline on both datasets; the frontier stays small because most recursion paths terminate quickly at exogenous turns and structural tool-result $\to$ tool-invocation hops incur no model call.
Recursive AttnTrace is the most expensive method because it pays the $B{=}30$ bootstrap cost at every frontier node; recursive \avgattn{} and AT2 are within 10\% of each other in absolute wall time.

\paragraph{Control backends.}
The two control backends of Table~\ref{tab:semantic-ablation} bracket the cost range of the paper.
Both are run on all four attribution models, over 688 ConFETTI targets and both TauBench domains; the per-target timings below are measured on \texttt{Qwen2.5-7B-Instruct}.
Semantic Retrieval is the cheapest scorer we evaluate: one MiniLM embedding pass and no forward pass through the attribution model, at 0.04--0.05\,s per flat target and 0.16\,s per recursive target.
ContextCite is the most expensive by two orders of magnitude, as each attribution call costs 64 ablation forward passes: 9.5\,s per flat target on ConFETTI and 25.5\,s on TauBench, rising to 23.3\,s and 70.0\,s for the recursive variants.
That is the same 2.5--3$\times$ recursive overhead the attention backends pay, but on a far larger base.
Its distribution is strongly right-skewed---medians of 16.2\,s and 52.4\,s against worst cases of 207\,s and 817\,s---because cost grows with both conversation length and DAG size.
For ContextCite, peak memory rather than time is the binding constraint: the ablation forward pass materialises $[\textit{batch} \times \textit{seq\_len} \times |V|]$ logits, so we cap the effective batch at $\textit{batch} \cdot \textit{seq\_len} \leq 32{,}768$ tokens and retry at a quarter batch size on OOM; mask seeds are deterministic, so a retry reproduces identical scores.
Flat and flat+structural variants cost the same to within 0.01\,s, since the structural edge adds no attribution call.

\paragraph{Total compute.}
The full set of results in Tables~\ref{tab:main_results} and~\ref{tab:appendix_main_results} requires four models
$\times$ six methods $\times$ ($493 + 3{,}099$) targets $= 86{,}208$ attribution
runs. On the hardware above this corresponds to approximately 117 GPU-hours of
compute (${\approx}9$\,h on ConFETTI, ${\approx}108$\,h on TauBench; summed
from per-target \texttt{timings} fields in output \texttt{meta.json} files),
plus a one-time labeling cost in GPT-5.4 inference for the first annotation
pipeline (109 ConFETTI conversations---79 used for evaluation, 30 held out for
$\alpha$ selection---and 576 TauBench simulations). The 117 GPU-hour total reflects the main evaluation runs only.
Additional attribution compute was incurred for the $\alpha$ sweep on the 30
held-out ConFETTI conversations (Section~\ref{sec:appendix-alpha}) and the
source granularity ablation on 10-conversation subsets per benchmark
(Section~\ref{sec:appendix-granularity}); the branching-factor ablation
(Section~\ref{sec:appendix-branching-ablation}) added no further cost, as it
re-analyzed stored graphs without additional attribution runs. The
control-backend runs add a further ${\approx}38$ GPU-hours, of which all but
under two are ContextCite.
No hyperparameter was tuned separately per method: $\alpha{=}0.85$, sentence
granularity, and $k{=}3$ are applied uniformly across all backends and
benchmarks.

\section{Ablation Studies}
\label{sec:appendix-ablations}

All ablations use the hyperparameters described in Appendix~\ref{sec:appendix-impl} unless stated otherwise.

\subsection{Recursion Depth Ablation}
\label{sec:appendix-depth-ablation}

We vary the maximum provenance depth $d_{\max}$ swept over $\{1, 2, 4, 8, 12\}$ for ConFETTI and $\{1, 2, 4, 8, 12, 16\}$ for TauBench, using Qwen2.5-7B-Instruct. Both sweeps run on the held-out slices (190 ConFETTI targets, 187 TauBench targets), not on the evaluation sets.
$d_{\max}$ is measured in DAG hops from the target node, counting every edge including structural tool-result $\to$ tool-invocation hops, so it is directly comparable to the provenance depth of gold annotations (Table~\ref{tab:dataset_overview}).
$d_{\max}{=}1$ retains only the target's direct parents, equivalent to flat attribution.

\begin{figure*}[t]
  \centering
  \includegraphics[width=\linewidth]{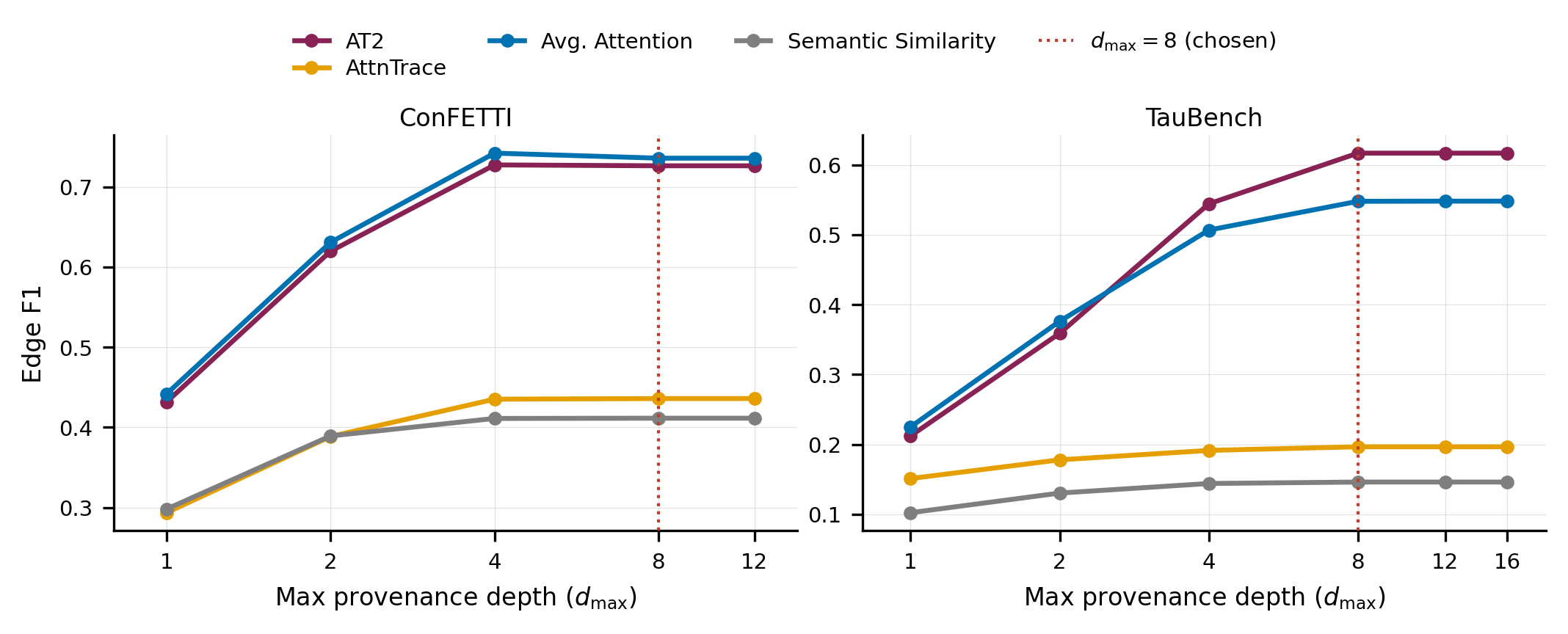}
  \caption{\textbf{Recursion depth ablation} (Qwen2.5-7B-Instruct; ConFETTI, left; TauBench, right).
  Edge F1 as a function of maximum provenance depth $d_{\max}$, measured in DAG hops (including structural tool-result $\to$ invocation edges).
  ConFETTI plateaus by $d_{\max}{=}4$ (average gold depth~3.6); TauBench by $d_{\max}{=}8$ (average gold depth~6.9). Both panels are computed on the held-out slices.
  Plateau depth tracks the gold depth distribution on each dataset.}
  \label{fig:recursion_depth}
\end{figure*}

Results are shown in Figure~\ref{fig:recursion_depth}.
On ConFETTI (average gold depth~3.6), \avgattn{} and AT2 rise sharply from $d_{\max}{=}1$ to $d_{\max}{=}4$ and plateau thereafter; $d_{\max}{=}4$ and $d_{\max}{=}8$ differ by under $0.01$ edge F1, so the plateau rather than its exact onset is the robust observation.
The plateau at $d_{\max}{=}4$ aligns directly with the dataset's average provenance depth: once the tracer can reach nodes four hops from the target, it has covered the majority of gold paths.

On TauBench (average gold depth~6.9), the curve saturates later, around $d_{\max}{=}8$, again tracking the deeper average gold structure.
The two datasets thus show different plateau points that match their respective depth distributions, validating that the ablation reflects genuine coverage of the gold provenance rather than a computational ceiling.

AttnTrace follows the same shape but at lower absolute values on both datasets, consistent with its weaker base-scorer performance in the main results.
The default $d_{\max}{=}8$ used in the main experiments corresponds to coverage well above the plateau on ConFETTI and at the plateau on TauBench, ensuring no gold paths are truncated.

\subsection{Pruning Threshold Ablation}
\label{sec:appendix-alpha}

\begin{figure}[h]
  \centering
  \includegraphics[width=\columnwidth]{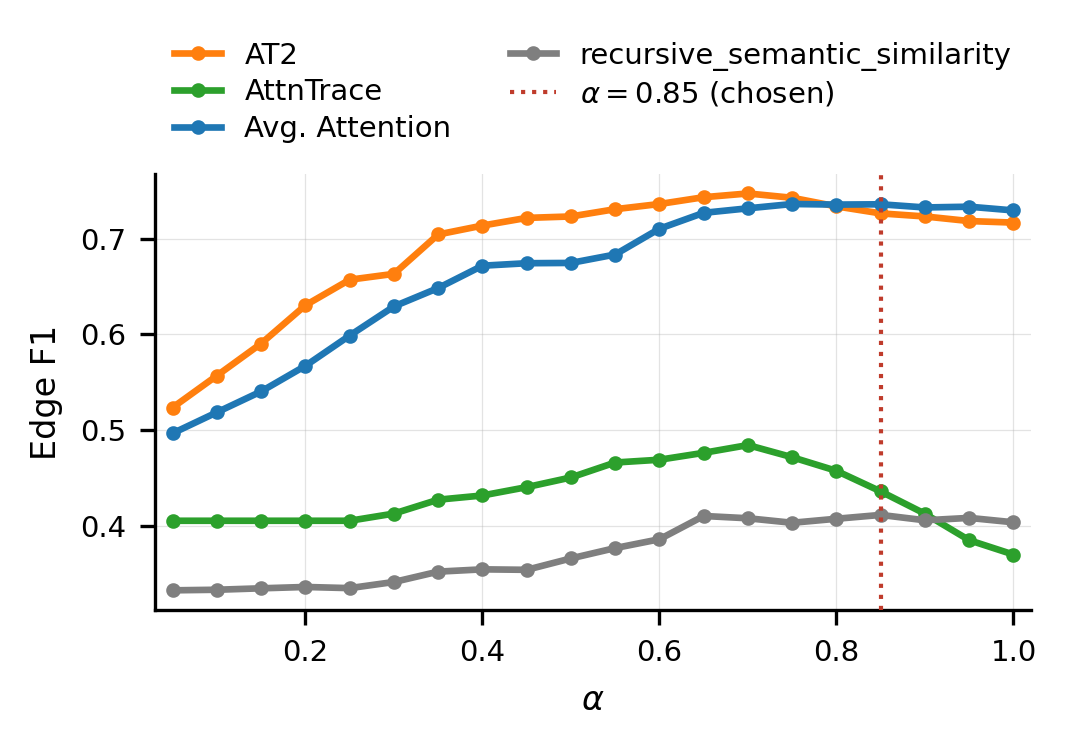}
  \caption{\textbf{Pruning threshold ablation} (Qwen2.5-7B-Instruct, held-out 30-conversation ConFETTI split, 190 targets).
    Edge F1 as a function of the relative-threshold extraction parameter $\alpha$ (Equation~\ref{eq:rel-threshold}), which controls how aggressively low-scoring edges are pruned from the predicted provenance graph before evaluation.
    Dashed line marks the chosen value $\alpha=0.85$; edge F1 varies by under 0.025 across $\alpha \in [0.65, 0.90]$ for both backends.}
  \label{fig:alpha-sweep}
\end{figure}

The pruning threshold (relative-threshold extraction) parameter $\alpha \in [0, 1]$ (Equation~\ref{eq:rel-threshold}) controls how aggressively edges are pruned from the predicted provenance graph before scoring.
We fix $\alpha{=}0.85$ and verify on the held-out 30-conversation ConFETTI split (190 targets, disjoint from the 79-conversation evaluation set). The sweep is run on ConFETTI only; no TauBench data was used to select $\alpha$.
As shown in Figure~\ref{fig:alpha-sweep}, $\alpha{=}0.85$ is optimal for \avgattn{} (0.736 edge F1); AT2 peaks slightly earlier at $\alpha{=}0.70$ (0.747) and is 0.021 lower at $\alpha{=}0.85$. Both vary by less than 0.025 edge F1 across $\alpha \in [0.65, 0.90]$, so the choice is insensitive within that band.
We therefore apply $\alpha = 0.85$ uniformly across all methods and both benchmarks.

\subsection{Branching Factor Ablation}
\label{sec:appendix-branching-ablation}

The per-step branching factor $k$ controls how many candidate parent spans the recursive tracer follows at each hop.
$k{=}1$ produces a greedy single chain (only the highest-scoring parent is recursed into at each step); $k{>}1$ allows multi-parent (DAG) recovery where multiple source spans jointly support a single intermediate.
We sweep $k \in \{1, 2, 3\}$ post-hoc on the stored graphs for the held-out 30-conversation ConFETTI split (190 targets), the same slice used for $\alpha$ and $d_{\max}$, applying relative-threshold extraction ($\alpha{=}0.85$) before top-$k$ selection.

\begin{figure}[h]
  \centering
  \includegraphics[width=\columnwidth]{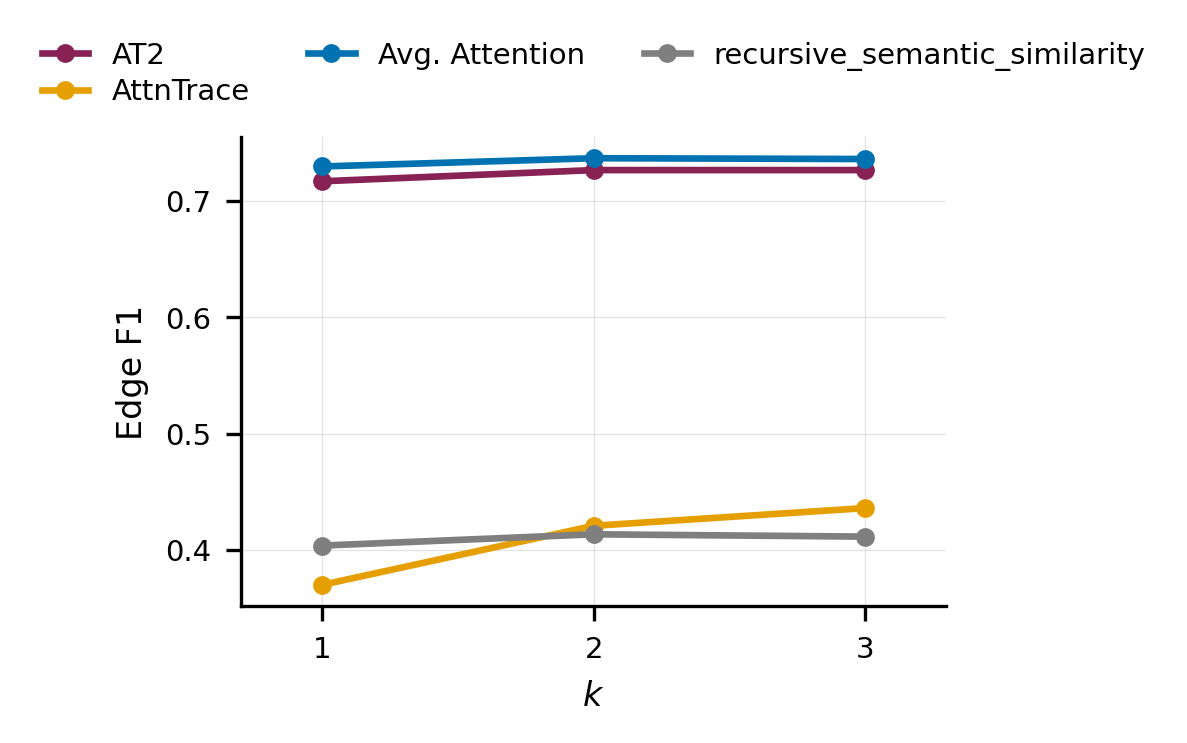}
  \caption{\textbf{Branching factor ablation} (Qwen2.5-7B-Instruct, held-out 30-conversation ConFETTI split, 190 targets, $\alpha{=}0.85$ relative-threshold pre-filtering).
  Edge F1 as a function of per-step branching factor $k \in \{1, 2, 3\}$.
  Results saturate at $k{=}2$; $k{=}3$ adds negligible further gain ($\Delta\text{F1}{<}0.001$).}
  \label{fig:branching-factor}
\end{figure}

Results for Qwen2.5-7B-Instruct on ConFETTI are shown in Figure~\ref{fig:branching-factor}.
\avgattn{} and AT2 each gain approximately $+0.01$ edge F1 from $k{=}1$ to $k{=}2$ ($0.729 \to 0.736$ and $0.717 \to 0.726$ respectively, as multi-parent gold edges become recoverable), and plateau exactly at $k{=}3$ ($\Delta\text{F1}{<}0.001$).
AttnTrace shows a slightly larger gain ($0.370 \to 0.421$ from $k{=}1$ to $k{=}2$) because its bootstrap-sampled scores are noisier and the single top-scored parent is less reliably the best one, but it also saturates by $k{=}3$.
The plateau reflects that most gold DAG nodes have at most two high-confidence parents after $\alpha{=}0.85$ pruning; a third branch virtually never passes the threshold.

For comparison, we also evaluate raw top-$k$ extraction without alpha pre-filtering on the same slice: F1 degrades monotonically with $k$ (from 0.729 at $k{=}1$ to 0.490 at $k{=}3$ for \avgattn{}) as predicted graphs become denser than the gold and precision collapses.
This confirms that $\alpha$ is the precision-restoring mechanism; $k$ governs recall coverage of multi-parent structure conditional on that pruning.

We adopt $k{=}3$ as the default: it recovers the full gain over the greedy chain baseline and leaves a conservative margin for rare 3-parent convergences, at negligible additional computation (the visited-set deduplication ensures each (turn, span) pair is attributed at most once per DAG regardless of $k$).

\subsection{Structural Edge vs.\ Recursion}
\label{sec:appendix-struct-ablation}

\textsc{Tokengeist} makes two changes to a single-pass attributor: it follows a
deterministic structural edge from each tool-result turn to its invoking tool-call
turn (Section~\ref{sec:tokengeist}), and it recurses into assistant turns. Since
the structural edge needs no model call and could be added onto any flat method,
we separate the two contributions by evaluating an intermediate baseline,
Flat+Struct: single-pass attribution plus the structural edge, with no
recursion.

\begin{table}[h]
\centering
\small
\setlength{\tabcolsep}{4pt}
\begin{tabular}{lrrr|rrr}
\toprule
& \multicolumn{3}{c|}{\textit{ConFETTI}} & \multicolumn{3}{c}{\textit{TauBench}} \\
\cmidrule(lr){2-4}\cmidrule(lr){5-7}
Base & Flat & +Struct & \textsc{Tg} & Flat & +Struct & \textsc{Tg} \\
\midrule
\multicolumn{7}{l}{\textit{Edge F1}} \\
Attention & 38.4 & 56.7 & \textbf{70.3} & 18.8 & 30.4 & \textbf{52.5} \\
AT2       & 40.2 & 58.5 & \textbf{72.4} & 18.1 & 28.7 & \textbf{54.5} \\
AttnTrace & 24.5 & 34.2 & \textbf{42.5} & 11.8 & 14.6 & \textbf{20.8} \\
\midrule
\multicolumn{7}{l}{\textit{Source recall}} \\
Attention & 33.7 & 52.0 & \textbf{91.9} & 15.4 & 16.2 & \textbf{86.9} \\
AT2       & 39.1 & 56.0 & \textbf{92.5} & 19.9 & 20.9 & \textbf{86.2} \\
AttnTrace & 46.8 & 60.7 & \textbf{86.6} & 28.1 & 32.1 & \textbf{64.8} \\
\bottomrule
\end{tabular}
\caption{\textbf{Separating the structural edge from recursion.} Edge F1 and
source recall, mean over the four attribution models, on the evaluation splits
($n{=}493$ / $3{,}099$). Flat+Struct is single-pass attribution with the
deterministic tool-result $\to$ tool-invocation edge and no recursion.}
\label{tab:struct-ablation}
\end{table}

The ordering of the two contributions inverts between the benchmarks. On ConFETTI
the structural edge covers more of the distance from Flat to
\textsc{Tokengeist} than recursion does; on TauBench, whose gold provenance is
deeper (average depth 6.9 against 3.6), recursion covers roughly twice as much as
the structural edge. This is the pattern the method predicts: a single structural
hop recovers one level of tool indirection, which is often what a shallow
conversation needs, while longer chains require recursion to traverse intermediate
assistant turns. We report Flat+Struct separately to show 
the structural edge is a genuine and cheap addition that any flat attributor could adopt.

The two metrics decompose differently.
On Edge F1 the structural rule is a substantial share of the gain---over half on
ConFETTI and roughly a third on TauBench. However, source recall shows 
its weakness: the structural edge adds 13.9--18.3 points on
ConFETTI and only 0.8--4.0 on TauBench, against 25.9--39.9 and 32.7--70.7 for
recursion's gains.

\subsection{Attribution Signal vs.\ Provenance Depth}
\label{sec:appendix-signal-depth}

We measure the \textit{signal coverage} of gold nodes (the fraction receiving any non-zero attribution score) as a function of (a) their provenance depth in the gold DAG and (b) their absolute turn distance from the target.
Depth is computed by BFS from the target along \textsc{ground\_truth\_deps} edges; invocation turns (assistant tool-call turns linked to their results via structural edges) appear at even depths~2, 4, \ldots.
These experiments use TauBench with Qwen2.5-7B-Instruct and all six methods.

\begin{figure*}[t]
  \centering
  \includegraphics[width=\linewidth]{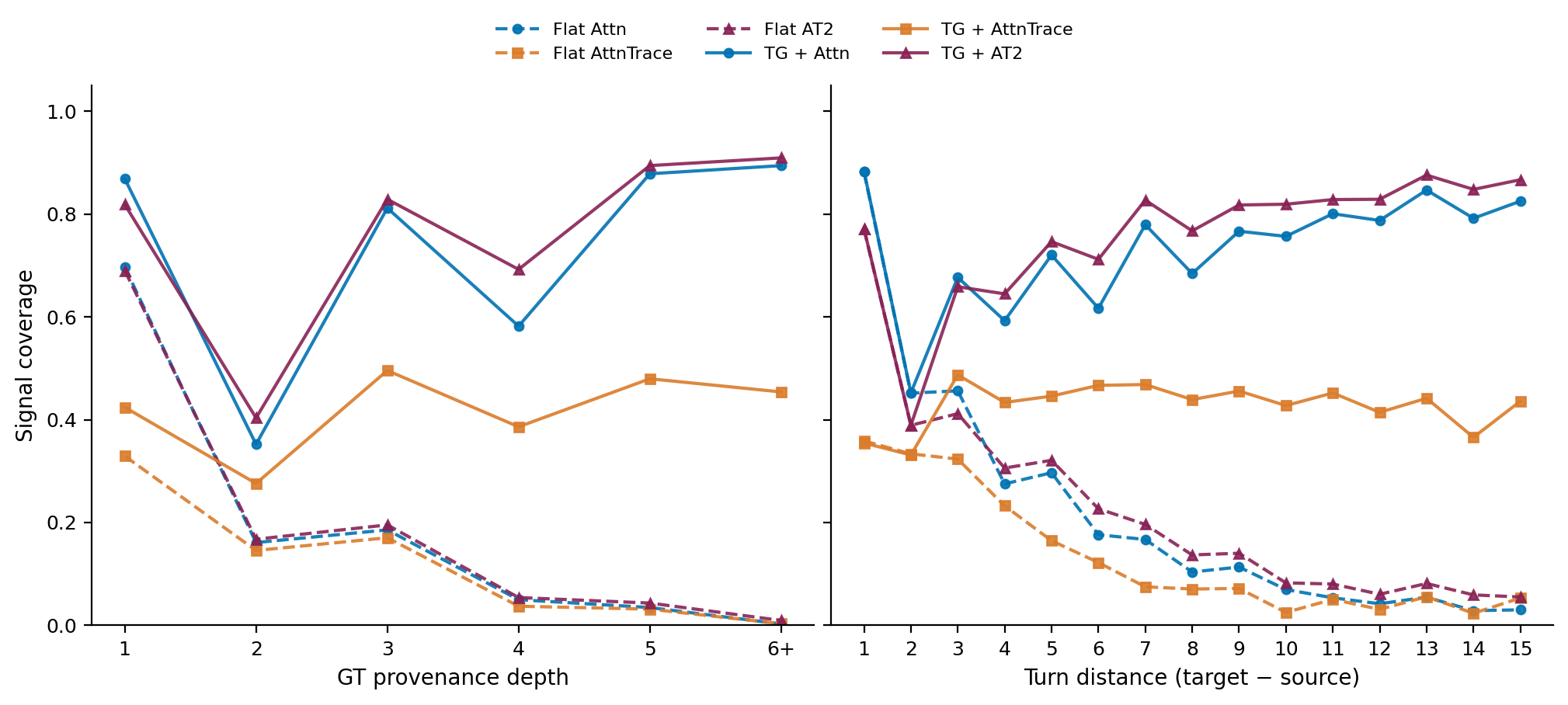}
  \caption{%
    \textbf{Signal coverage of gold nodes vs.\ provenance depth (left) and turn distance (right)}
    (TauBench, Qwen-2.5-7B, $n{=}26{,}478$ gold-node records).
    Signal coverage = fraction of gold nodes whose turn receives any non-zero attribution score.
    Flat methods (dashed lines) collapse beyond depth~1 / distance~2;
    recursive \textsc{Tokengeist} variants (solid lines) sustain signal coverage above 75\% at all depths and distances tested.
    Even-depth dips in recursive curves reflect invocation turns propagated via structural (zero-signal) edges.
  }
  \label{fig:signal-vs-depth}
\end{figure*}

Figure~\ref{fig:signal-vs-depth} (left) shows signal coverage by depth.
All flat methods (dashed) cover roughly 70\% of depth-1 gold nodes but drop below 20\% at depth~2 and approach zero beyond depth~3.
Recursive \textsc{Tokengeist} variants (solid) maintain signal coverage above 80\% at odd depths~3, 5, 7 (exogenous turns: tool results, user messages) and above 50\% at even depths (invocation turns).
The even-depth dip reflects that invocation turns are connected via zero-cost structural edges that carry no attribution signal; they are still reached by the tracer but their soft score is zero unless the recursive call above them also assigns a direct non-structural edge.
\textsc{Tg}+Attention and \textsc{Tg}+AT2 reach ${\approx}90\%$ signal coverage at depth~${\geq}6$.

Figure~\ref{fig:signal-vs-depth} (right) shows signal coverage by turn distance.
The advantage of recursion emerges from distance~3 onward: flat signal coverage halves by distance~2 and falls below 10\% beyond distance~5, while recursive methods sustain coverage above 75\% out to distance~15 and beyond.

\paragraph{Depth versus distance.}
Depth and distance are correlated, so Figure~\ref{fig:signal-vs-depth} does not
by itself separate provenance collapse from positional
degradation~\citep{liu-etal-2024-lost}. Table~\ref{tab:depth-distance} holds
each fixed in turn. At the nearest distances, where position should not hurt,
flat recovery still collapses with depth: 65.0 to 0.0 on ConFETTI, 59.1 to 3.3
on TauBench. At fixed depth~1 it falls far less over the whole distance range,
65.0 to 31.8. This isolates the effect of recursion, because it moves no source, only the
number of hops traversed. 
\textsc{Tokengeist} stays between 86.8 and 95.5 on ConFETTI at every depth and
distance.

\begin{table}[h]
\centering
\small
\setlength{\tabcolsep}{4pt}
\begin{tabular}{lrrr|rrr}
\toprule
& \multicolumn{3}{c|}{\textit{ConFETTI}} & \multicolumn{3}{c}{\textit{TauBench}} \\
\cmidrule(lr){2-4}\cmidrule(lr){5-7}
& \multicolumn{6}{c}{\footnotesize turn distance} \\
Depth & 1--4 & 5--9 & 10+ & 1--4 & 5--9 & 10+ \\
\midrule
\multicolumn{7}{l}{\textit{Flat Attention}} \\
1    & 65.0 & 39.7 & 31.8 & 59.1 & 26.2 & 18.6 \\
2--3 & 29.9 & 20.6 & 16.4 & 21.0 & 10.5 &  3.7 \\
4+   &  0.0 &  7.7 &  6.2 &  3.3 &  6.0 &  0.4 \\
\midrule
\multicolumn{7}{l}{\textit{\textsc{Tg} + Attention}} \\
1    & \textbf{90.4} & \textbf{92.6} & \textbf{95.5} & \textbf{76.0} & \textbf{86.1} & \textbf{92.8} \\
2--3 & \textbf{91.4} & \textbf{88.2} & \textbf{91.4} & \textbf{66.7} & \textbf{73.0} & \textbf{91.2} \\
4+   & \textbf{90.0} & \textbf{86.8} & \textbf{93.5} &  6.7 & \textbf{87.6} & \textbf{95.6} \\
\bottomrule
\end{tabular}
\caption{\textbf{Gold-source recovery by provenance depth and turn distance.}
Percentage of gold sources recovered, mean over the four attribution models, on
the evaluation splits of Table~\ref{tab:main_results}. Depth is the shortest hop
count from the target over \textsc{ground\_truth\_deps}; distance is the number
of turns between the source and the target. Sources per cell, per model, by row:
Depth~4+ within four turns forces one gold hop per intervening turn, so that cell
($n{=}15$) is too small to read.}
\label{tab:depth-distance}
\end{table}

\subsection{Source Granularity Ablation}
\label{sec:appendix-granularity}

\begin{figure*}[h]
  \centering
  \includegraphics[width=\linewidth]{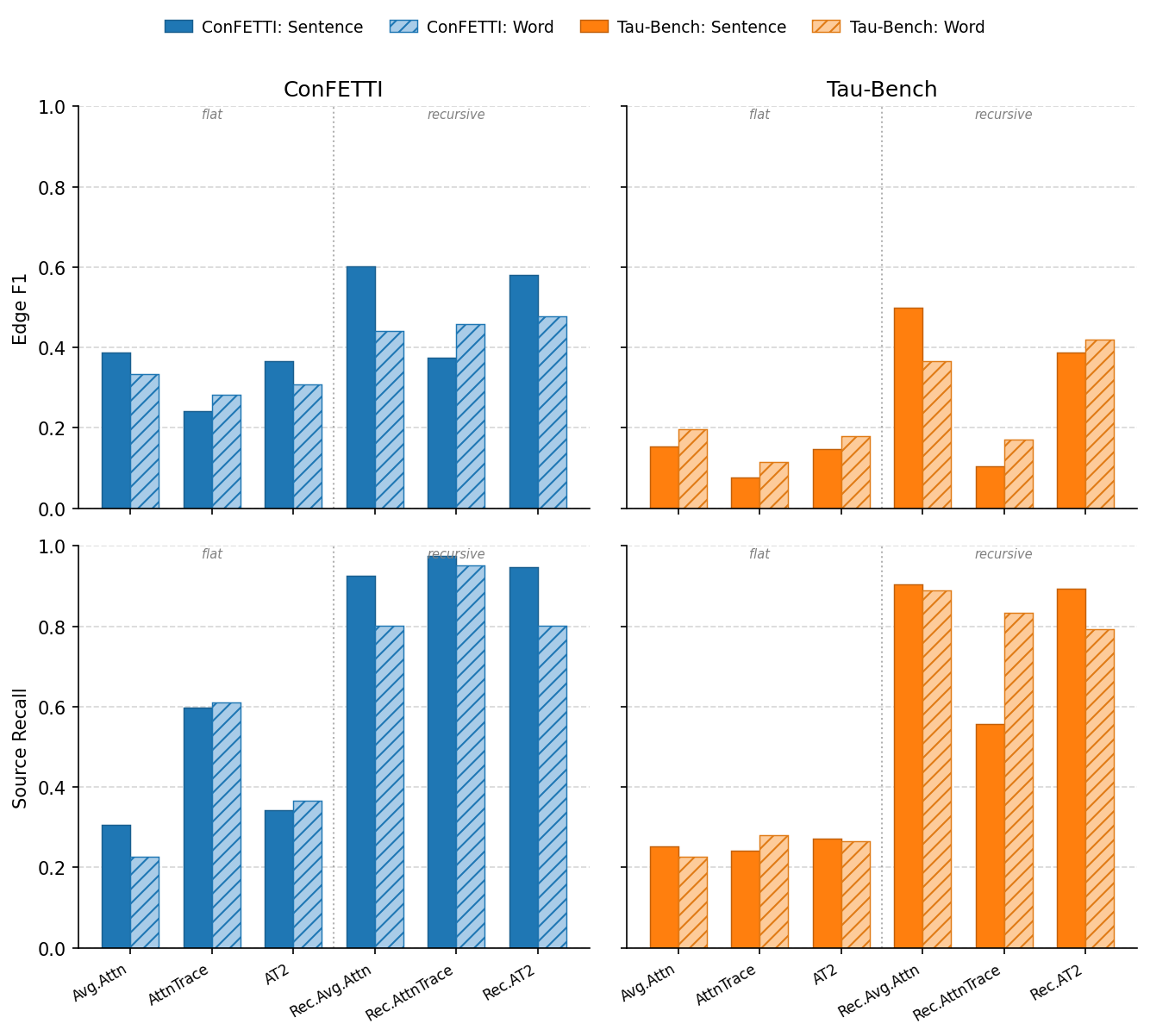}
  \caption{\textbf{Source granularity ablation} (Llama-3.1-8B-Instruct, 10-conversation subset).
  Edge F1 (top) and source recall (bottom) for sentence- vs.\ word-level segmentation across all backends on ConFETTI (left) and TauBench (right).
  Sentence is better for \avgattn{} and AT2; word is better for AttnTrace due to its bootstrap segment-sampling design, but at much higher runtime cost.}
  \label{fig:granularity-ablation}
\end{figure*}

We compare two source segmentation granularities, \textit{sentence-level} (default) and \textit{word-level}, across all three attribution backends on 10 ConFETTI conversations (62 targets) and 10 TauBench simulations (60 targets) using Llama-3.1-8B-Instruct.
Results after relative-threshold extraction ($\alpha = 0.85$) are shown in Figure~\ref{fig:granularity-ablation}.

The effect of granularity is method-dependent.
For \avgattn{}, sentence granularity outperforms word on the recursive variants across both datasets: on ConFETTI, edge F1 is $+0.16$ higher and source recall $+0.13$ higher; on TauBench, edge F1 is $+0.13$ higher.
AT2 follows a similar pattern on ConFETTI ($+0.11$ edge F1, $+0.15$ source recall) but the advantage is smaller and mixed on TauBench.
This is consistent with the well-documented tendency of attention weights to concentrate on punctuation and special tokens~\citep{xiao2024efficient}: scoring sentence-length spans as atomic units averages out these spikes, whereas word-level segments are small enough that a single high-attention function word can dominate the score for that segment.

AttnTrace shows the opposite pattern: word granularity is consistently better, particularly on TauBench where recursive AttnTrace gains $+0.07$ edge F1 and $+0.28$ source recall with word segmentation.
AttnTrace scores segments by bootstrap subsampling of context groups, so smaller word-level segments give the sampler finer-grained control over which content is included or excluded per iteration.
However, word-level segmentation substantially increases runtime for AttnTrace: the number of bootstrap iterations scales with the number of segments, so word-level inputs require far more forward passes than sentence-level inputs, making it impractical at scale.

Since sentence granularity is optimal for both \avgattn{} and AT2 on both benchmarks, and word-level segmentation is impractical for AttnTrace at scale, we adopt sentence-level granularity as the fixed default for all experiments.

\section{Additional Attribution Backends}
\label{sec:appendix-backends}

This section details attribution methods that do not read attention weights,
semantic similarity, and counterfactual ablation.
Together with Flat+Struct (Appendix~\ref{sec:appendix-struct-ablation}), these
are the three controls that isolate the contribution of recursion
(Section~\ref{sec:experiments}).

\subsection{Semantic Retrieval Attribution Backend}
\label{sec:appendix-semantic}

Semantic scores each candidate source span by cosine similarity between
\texttt{all-MiniLM-L6-v2} embeddings of that span and the target span. We evaluate it
in both settings: a single pass (Flat Semantic)
and the recursive \textsc{Tokengeist} traversal (\textsc{Tg}~+~Semantic). Because the
scorer sees only surface text, the distance between those two rows is
attributable to the traversal alone, with the attribution signal held fixed.

\paragraph{Flat Semantic Retrieval.}
A single pass of Semantic is plain embedding retrieval over the whole
prior context. While it is a weak attributor overall, it recovers source recall
better than flat attention (59.3 source recall against 35.1 on ConFETTI, 42.7 against
17.9 on TauBench) but is markedly \emph{worse} at finding intermediate turns
(26.7 edge F1 against 39.5, 11.5 against 22.5). Attention localizes structure
but cannot reach back; retrieval reaches back but mis-assigns the structure.
This is the same pattern the main results show between Flat Attention and
Flat AttnTrace (Table~\ref{tab:main_results}), in a sharper form.

\paragraph{Recursive Semantic Retrieval.}
In the recursive setting, \textsc{Tg}~+~Semantic follows the same top-$k$ selection, 
pruning threshold $\alpha$, $d_{\max}$, and deterministic structural edge 
as \textsc{Tg}~+~Attention. Recursion raises source recall from 59.3 to 81.9 on
ConFETTI and from 42.7 to 82.8 on TauBench, gains of $22.6$ and $40.1$, with a
smaller edge-F1 gain ($10.7$ and $3.5$). Because the scorer reads no model
internals these figures barely move with the attribution model---the
source-recall gain is $22.5$ on ConFETTI and $39.3$--$40.5$ on TauBench across
all four---and it is positive for every model on both benchmarks, clearly
capturing the contribution of recursion.

\begin{table}[t]
\centering\small\setlength{\tabcolsep}{3pt}
\begin{tabular}{llrr|rr}
\toprule
& & \multicolumn{2}{c|}{\textit{ConFETTI}} & \multicolumn{2}{c}{\textit{TauBench}} \\
\cmidrule(lr){3-4}\cmidrule(lr){5-6}
Scorer & & Edge & Span & Edge & Span \\
\midrule
Attention    & Flat           & 38.4 & 12.0 & 18.8 & 4.9 \\
             & +\,\textsc{Tg} & 70.3 & 46.9 & 52.5 & 35.6 \\
\midrule
AT2          & Flat           & 40.2 & 13.6 & 18.1 & 5.6 \\
             & +\,\textsc{Tg} & 72.4 & 47.2 & 54.5 & 36.8 \\
\midrule
AttnTrace    & Flat           & 24.5 & 13.8 & 11.9 & 5.3 \\
             & +\,\textsc{Tg} & 42.5 & 32.2 & 20.8 & 14.4 \\
\midrule
Semantic     & Flat           & 26.7 & 10.6 & 10.3 & 5.7 \\
             & +\,\textsc{Tg} & 37.4 & 30.0 & 14.7 & 10.8 \\
\midrule
ContextCite  & Flat           & 38.1 & 8.6 & 17.1 & 4.6 \\
             & +\,\textsc{Tg} & 65.8 & 40.3 & 48.6 & 30.4 \\
\bottomrule
\end{tabular}
\caption{\textbf{Recursion lifts every base scorer.}
Edge and Span F1 for each scorer applied flat and under the \textsc{Tokengeist}
traversal; each cell is the mean over the four attribution models, with the
per-model range plotted in Figure~\ref{fig:recursion-gain}. Scored on
the same evaluation split and per-metric target sets as
Tables~\ref{tab:main_results} and~\ref{tab:appendix_main_results}
($n{=}493$ ConFETTI, $n{=}3{,}081$ TauBench, pooling the airline and retail
domains).
Per-model ContextCite figures on all four metrics are in
Table~\ref{tab:contextcite-permodel}.}
\label{tab:semantic-ablation}
\end{table}

\begin{table*}[t]
\centering\small\setlength{\tabcolsep}{4pt}
\begin{tabular}{llrrrr|rrrr}
\toprule
& & \multicolumn{4}{c|}{\textit{ConFETTI}} & \multicolumn{4}{c}{\textit{TauBench} (airline\,+\,retail)} \\
\cmidrule(lr){3-6}\cmidrule(lr){7-10}
Model & & Edge F1 & Node F1 & Src.\ Rec.\ & Span F1 & Edge F1 & Node F1 & Src.\ Rec.\ & Span F1 \\
\midrule
Qwen2.5-7B & Flat           & 38.6 & 44.4 & 41.0 & 8.7 & 17.9 & 23.5 & 22.9 & 4.8 \\
                             & +\,\textsc{Tg} & 67.0 & 77.1 & 84.5 & 40.3 & 46.8 & 61.2 & 76.0 & 27.5 \\
\midrule
Llama-3.1-8B & Flat           & 38.3 & 44.5 & 38.5 & 8.8 & 16.1 & 21.3 & 27.7 & 4.7 \\
                             & +\,\textsc{Tg} & 65.1 & 76.4 & 83.2 & 40.6 & 38.1 & 54.0 & 78.9 & 27.6 \\
\midrule
Phi-4-mini & Flat           & 37.7 & 43.3 & 38.8 & 8.2 & 17.7 & 22.8 & 22.0 & 4.5 \\
                             & +\,\textsc{Tg} & 65.4 & 76.3 & 83.6 & 40.3 & 55.3 & 65.0 & 80.3 & 33.3 \\
\midrule
DeepSeek-R1-Qwen-7B & Flat           & 37.8 & 43.9 & 38.1 & 8.5 & 16.7 & 22.1 & 23.3 & 4.6 \\
                             & +\,\textsc{Tg} & 65.7 & 75.8 & 84.4 & 40.0 & 54.0 & 64.5 & 77.5 & 33.3 \\
\bottomrule
\end{tabular}
\caption{\textbf{ContextCite per attribution model.} The counterfactual
scorer applied flat and under the \textsc{Tokengeist} traversal, on the same
evaluation split and per-metric target sets as Tables~\ref{tab:main_results}
and~\ref{tab:appendix_main_results}. The recursive traversal lifts every metric
for every model.}
\label{tab:contextcite-permodel}
\end{table*}

\begin{figure*}[t]
\centering
\includegraphics[width=0.86\textwidth]{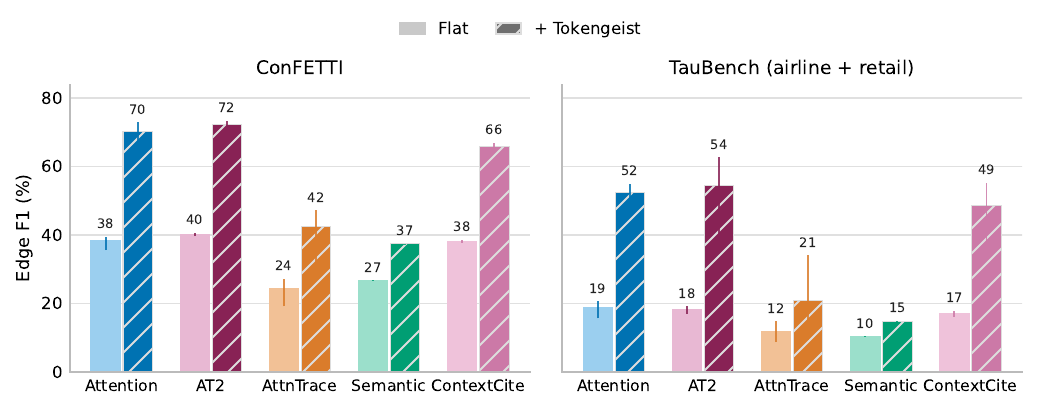}
\caption{\textbf{Every base scorer gains from the recursive traversal.}
Edge F1 for each base scorer applied flat (light) and under the
\textsc{Tokengeist} traversal (saturated, hatched), mean over the four
attribution models; the vertical rule on each bar spans the per-model
minimum and maximum. The gain holds consistently across all four models for the counterfactual scorer
(ContextCite), model-free scorer (Semantic), as well as attention-based scorers
(Attention, AT2, AttnTrace).}
\label{fig:recursion-gain}
\end{figure*}

\begin{figure}[h]
\centering
\includegraphics[width=\columnwidth]{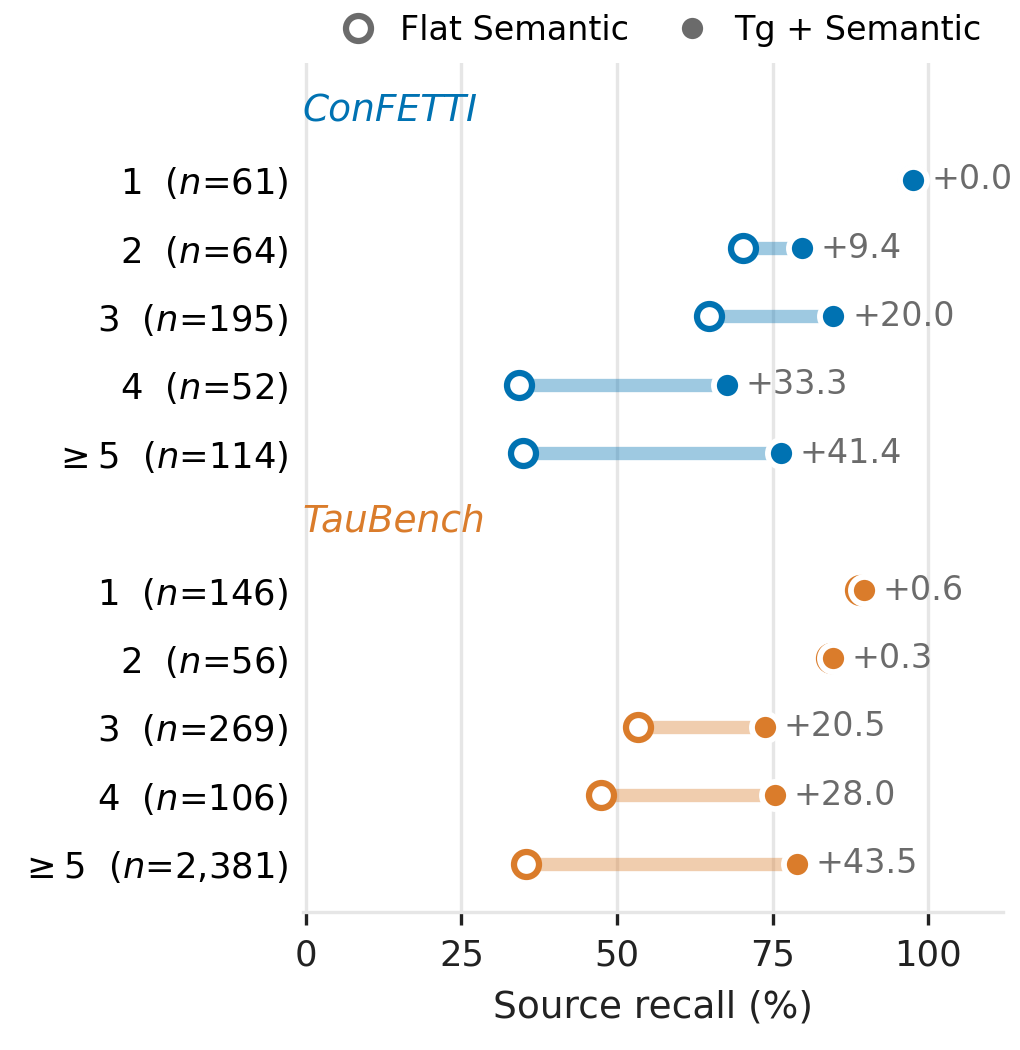}
\caption{\textbf{Where recursion helps.} Source recall of Flat Semantic and
\textsc{Tg}~+~Semantic by gold chain depth, mean over the four attribution
models. Each segment is the recursion gain at that depth. Depth is the longest
target-to-source path in the gold DAG, as in Table~\ref{tab:dataset_overview};
$n$ is targets per model.}
\label{fig:semantic-depth}
\end{figure}

The gain is concentrated in deep chains (Figure~\ref{fig:semantic-depth}).
The advantage of recursion grows steeply at depth~3 onward, reaching $+41.4$ and $+43.5$ at
depth~$\geq5$ on both benchmarks. The larger overall
average on TauBench therefore reflects its heavier depth distribution.

\paragraph{What this bounds.}
Against the attention-based scorers, \textsc{Tg}~+~Semantic recovers most of the
source recall---closing roughly four-fifths of the Flat Attention to
\textsc{Tg}~+~Attention gap on ConFETTI and about seven-eighths on TauBench---but
little of the edge F1, reaching 37.4 and 15.0 against 73.0 and 52.7. 
Although \textsc{Tg}~+~Semantic scores well on source recall, it 
cannot recover the structure that attention provides.

\subsection{Counterfactual Attribution Backend}
\label{sec:appendix-contextcite}

ContextCite~\citep{cohen2024contextcite} fits a sparse linear
surrogate predicting the target's logit-probability from a binary vector
indicating which source spans are present in the context. The surrogate's
weight for a source is that source's attribution score. A high score therefore
means the target's probability \emph{drops} when the source is ablated, not
that the model attended to it. We ablate a source by zeroing its attention mask
in place rather than deleting its text and re-rendering the prompt, which holds
the target's token range fixed across ablations and keeps the scores
commensurable with the other backends; the surrogate is fit on 64 ablations per
call with keep probability $0.5$ and L1 strength $0.01$.

Instead of attention-mediated influence with, ContextCite measures the causal effect of a source on the target's
counterfactual necessity. Table~\ref{tab:semantic-ablation} reports both
alongside the attention-based scorers, averaged over the four attribution models
and pooling both TauBench domains; Table~\ref{tab:contextcite-permodel} breaks
ContextCite out per model.

\paragraph{Flat ContextCite.}
On ConFETTI with Qwen2.5-7B-Instruct, a single counterfactual pass finds DAG
structure comparatively to attention but reaches exogenous turns worse than
other scorers: 41.6 source
recall against 56.0 for AttnTrace and 59.3 for semantic retrieval, and the
lowest span F1 of the five at 8.7. This inverts the semantic retrieval pattern,
which reaches back well but mis-assigns structure, and likely follows from the
surrogate's sparsity. Sources are sentences, and a ConFETTI conversation carries a median
of 77 of them against 64 ablations, so the L1 fit is underdetermined by
construction and returns a small set of non-zero weights. 

\paragraph{Recursive ContextCite.}
Recursion significantly lifts every metric; on Qwen2.5-7B-Instruct, edge F1 rises
from 38.6 to 67.0, node F1 from 44.4 to 77.1, source recall from 41.6 to 85.7,
and span F1 from 8.7 to 40.3, and the lift is present for all four attribution
models on both datasets. Decomposing the gain against the
flat+structural variant (Section~\ref{sec:appendix-struct-ablation}), which
reaches 55.9 edge F1 on these targets, the deterministic tool edge accounts for
$17.3$ of the $28.4$ points and recursion into assistant turns for the
remaining $11.1$ --- a split comparable to Attention's $18.6$ and $14.9$ out of
$33.5$. Thus, the recursion component is not an artifact of the hard-coded structural
rule under either scorer.

\paragraph{What this bounds.}
The compositional structure the recursive traversal recovers is therefore
not specific to attention, as it is evident for a counterfactual
scorer as well, and across all four attribution models. Attention-based
attributors still outperform ContextCite in within-turn localization (span F1
45.7--48.2 against 40.0--40.6 on ConFETTI, 34.0--36.4 against 27.5--33.3 on
TauBench) and in reaching the originating event (source recall 89.5--94.1
against 83.2--84.5, and 82.4--89.0 against 76.0--80.3), in every case for every
model. The Edge F1 ordering, however, does not survive the wider scope: on
TauBench ContextCite overtakes attention on two of the four models
(55.3 against 52.3 on Phi-4-mini, 54.0 against 50.0 on
DeepSeek-R1-Distill-Qwen-7B), and on ConFETTI the two tie at 65.7 on
DeepSeek-R1. Recovering the dependency structure is thus not an advantage of
attention over counterfactual necessity; localizing within a turn and reaching
back to the originating event are. Furthermore, ContextCite costs significantly 
more compute; each ContextCite call costs 64 forward passes over the conversation, 
against one call for the attention-aggregation backends.

\section{MTCABench Details}
\label{sec:appendix-dataset}

\subsection{Dataset Statistics}

\paragraph{Dependency type counts.}
Table~\ref{tab:dep_tags} gives per-tag counts for both splits.
Base tags (\textit{direct}/\textit{chained}) are mutually exclusive; additional tags can co-occur with either base tag.

\begin{table}[h]
\centering
\small
\begin{tabular}{lrrrr}
\toprule
 & \multicolumn{2}{c}{\textbf{ConFETTI}}
 & \multicolumn{2}{c}{\textbf{TauBench}} \\
\cmidrule(lr){2-3}\cmidrule(lr){4-5}
Tag & Count & \% & Count & \% \\
\midrule
\multicolumn{5}{l}{\textit{Base tag}} \\
\quad direct   & 104 & 15.2 &   267 &  8.1 \\
\quad chained  & 579 & 84.8 & 3{,}020 & 91.9 \\
\midrule
\multicolumn{5}{l}{\textit{Additional tag}} \\
\quad multi\_source  & 206 & 30.2 & 2{,}391 & 72.7 \\
\quad hallucination  &  58 &  8.5 &    50 &  1.5 \\
\bottomrule
\end{tabular}
\caption{\textbf{Dependency-type tag distribution in \textsc{MTCABench}.}
Percentages are out of every released target in each split: ConFETTI ($n$=683), TauBench ($n$=3,287), which includes the held-out slices used for hyperparameter selection.
A single target may carry multiple additional tags but only one base tag.
The base tags and \textit{multi\_source} are derived from the released graph (Appendix~\ref{sec:appendix-annotation-quality}); \textit{hallucination} and \textit{tool\_call} are annotated.}
\label{tab:dep_tags}
\end{table}

\begin{figure}[h]
  \centering
  \includegraphics[width=\linewidth]{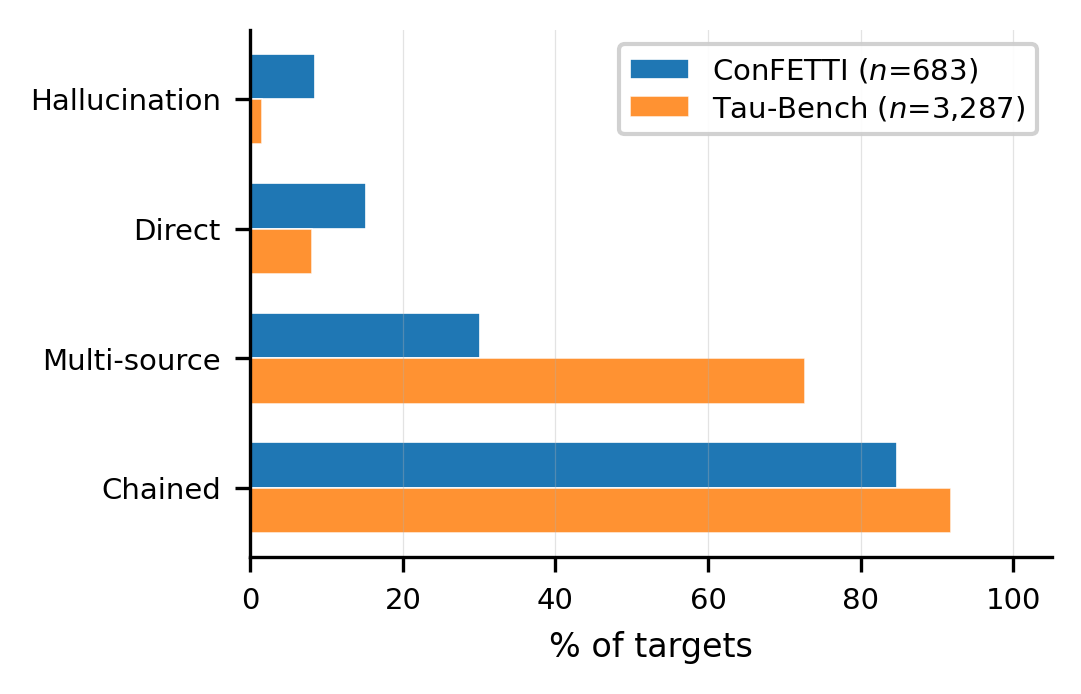}
  \caption{\textbf{Dependency-type tag distribution in \textsc{MTCABench}.}
  Bars show the percentage of targets in each split carrying each tag.}
  \label{fig:dep_tag_distribution}
\end{figure}

\paragraph{Dependency type distribution.}
Figure~\ref{fig:dep_tag_distribution} shows the percentage of targets carrying each dependency tag, for both splits.

\paragraph{Conversation length statistics.}
Table~\ref{tab:conv_stats} reports conversation length for both splits.
TauBench token counts are exact values from the \texttt{usage.prompt\_tokens} field recorded during simulation; ConFETTI token counts are character-based estimates ($\approx$chars\,/\,4).
The longer contexts in TauBench reflect multi-step agentic tool-call chains, which also drive the higher average provenance depth (Table~\ref{tab:dataset_overview}).

\begin{table}[h]
\centering
\small
\begin{tabular}{lrr}
\toprule
 & \textbf{ConFETTI} & \textbf{TauBench} \\
\midrule
Turns per conv: mean & 16.8 & 27.3 \\
Turns per conv: range & 9--35 & 14--55 \\
Input tokens: mean   & $\approx$2{,}143 & 7{,}082 \\
Input tokens: range  & $\approx$1{,}000--7{,}300 & 4{,}161--12{,}495 \\
\bottomrule
\end{tabular}
\caption{\textbf{Conversation length statistics.}
Input token counts for ConFETTI are character-based estimates; TauBench values are exact prompt-token counts from simulation logs.}
\label{tab:conv_stats}
\end{table}

The 576 TauBench simulations cover two domains: airline (129 simulations, 22.4\%) and retail (447 simulations, 77.6\%).

\paragraph{Domain coverage.}
Table~\ref{tab:domains} lists the 10 ConFETTI task families identified by semantic clustering and the 2 TauBench domains.

\begin{table*}[t]
\centering
\small
\begin{tabular}{llrl}
\toprule
\textbf{Split} & \textbf{Domain} & \textbf{Conv.} & \textbf{Representative APIs} \\
\midrule
\multirow{10}{*}{ConFETTI}
 & Accommodation \& Car Rental & 18 & BookHotel, CarRental \\
 & Lifestyle \& Local          & 16 & RestaurantSearch, Shopping, NewsSearch \\
 & Medical Services            & 15 & DoctorsAppointment, Telehealth \\
 & Insurance \& Benefits       & 14 & Insurance, DoctorsAppointment \\
 & Business Travel \& HR       & 12 & BookFlight, HRTimeOff, HRCompanyDirectory \\
 & Telecom \& Social           & 11 & Telecom, Discord \\
 & Banking                     &  7 & Banking \\
 & HR Operations               &  6 & HRCompanyDirectory, HRPayrollBenefits \\
 & Productivity                &  6 & MapMyFitness, GMail, Slack \\
 & Vacation Planning           &  4 & BookFlight, CarRental, Weather \\
\midrule
\multirow{2}{*}{TauBench}
 & Retail  & 447 & order management, product search, returns \\
 & Airline & 129 & flight booking, reservations, baggage \\
\bottomrule
\end{tabular}
\caption{\textbf{Task domain coverage in \textsc{MTCABench}.}
ConFETTI domains are derived from k-means clustering ($k{=}10$) of \texttt{all-MiniLM-L6-v2} embeddings of API call signatures; TauBench domains are provided by the dataset.
TauBench airline overlaps topically with ConFETTI travel clusters, giving 11 distinct task families in total.}
\label{tab:domains}
\end{table*}

\begin{figure}[h]
  \centering
  \includegraphics[width=\linewidth]{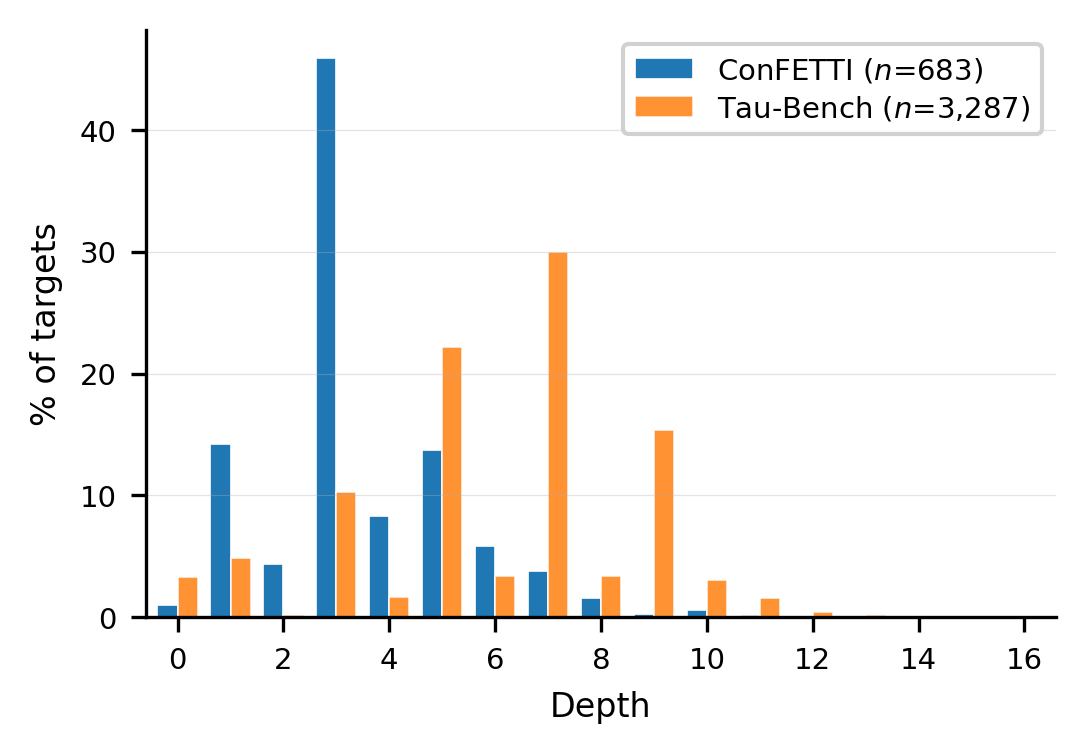}
  \caption{%
    Ground-truth provenance depth distribution across all 683 released ConFETTI targets and 3,287 released TauBench targets in \textsc{MTCABench}; the figure characterises the released benchmark, so it includes the held-out hyperparameter slices.
    Depth is the maximum number of edges on any root-to-leaf path in the provenance graph.
    ConFETTI peaks at depth~3 ($n$=306), while TauBench peaks at depth~7 ($n$=875),
    reflecting the longer tool-call chains in agentic simulations.
    Depths span 0--12 on ConFETTI and 0--17 on TauBench, covering ungrounded targets, simple single-hop cases, and deep multi-hop attribution scenarios.
  }
  \label{fig:depth_hist}
\end{figure}

\paragraph{Provenance depth distribution.}
Figure~\ref{fig:depth_hist} shows the full depth distribution for both splits.

\subsection{Annotation Prompts}
\label{sec:appendix-prompts}

Listing~\ref{lst:prompt-phase1} is the phase-1 prompt, which identifies targets and
runs once. Listing~\ref{lst:prompt-phase2} is the phase-2 prompt, used by three of
the four phase-2 annotators.
The phase-2 prompt is aligned with the review rubric of step~3b and hardened against
each defect class the initial \texttt{GPT-5.4} sweep found. 


\begin{figure*}[t]
\begin{lstlisting}[
  style=promptstyle,
  caption={\textbf{Phase 1: Target Identification Prompt.} Used to select 5--8 attribution target spans per conversation. \texttt{\{conversation\}} is replaced by the full serialised conversation.},
  label={lst:prompt-phase1}
]
You are selecting annotation targets for a multi-turn LLM conversation benchmark
on token attribution.

A good target span is:
- Specific text in an ASSISTANT turn that carries factual content (a value, date,
ID, location, amount, or a natural-language paraphrase of tool output). NEVER
target tool result turns or user turns, only assistant turns.
- Traceable back to at least one earlier turn through a clear dependency path.
- Varied across dep types: aim for a mix of direct (1-hop from source), chained
(multi-hop through tool calls or intermediary turns), and multi-source (target
requires information from two distinct originating turns)
- Look for hallucination targets - spans where the model used the wrong value,
traceable to a spurious prior turn rather than the correct source. For example: an
argument value copied from a previous tool call instead of read from the current
tool result. These are particularly valuable benchmark cases.

Aim for 5-8 targets. Spread them across different turns and different parts of the
conversation.
Prefer concrete specificity - an exact ID, date, or quoted value is better than a
vague phrase.

CONVERSATION: {conversation}
Return a JSON object with a single key "targets" containing the array, no other
text.
{
    "targets": [
      {
          "name": "<short snake_case identifier describing the target>",
          "target_turn_idx": <int, 0-based index of the turn containing the span>,
          "target_text": "<exact verbatim span copied from that turn>"
      }
    ]
}
\end{lstlisting}
\end{figure*}

\begin{figure*}[t]
\begin{lstlisting}[
  style=promptstyle,
  firstline=1,
  lastline=45,
  firstnumber=1,
  caption={\textbf{Phase 2: Provenance Annotation Prompt} (part 1 of 3). Used by the three annotators that share it. \texttt{\{conversation\}} and \texttt{\{targets\_json\}} are replaced at call time; \texttt{\{sim\_id\}} is the conversation identifier.},
  label={lst:prompt-phase2}
]

You are annotating a multi-turn conversation for a research benchmark on token attribution. The goal is a ground-truth dependency DAG capturing which prior turns 
causally produced a specific target span.

A conversation consists of user, assistant, and tool turns, indexed from 0. Tool call turns (assistant turns that invoke functions) and tool result turns are included and indexed.

DEFINITIONS
Target: a specific text span within a specific assistant turn whose provenance you trace.
Dependency: turn T depends on prior turn S if the specific content of S was causally necessary to produce T. A turn that is merely contextually relevant is not a dependency: include S only if, without it, T would be wrong, incomplete, or different.
Source node (leaf): a turn with no further dependencies, with depends_on: [].
Exogenous turn: a user, system, or tool-result turn -- where content enters the conversation.

EDGE ROLES -- what to include
- value: the parent supplies content the child uses, including content that arrives transformed (a date reformatted, "701" becoming a street_number field, a city name becoming a state code). ALWAYS include.
- selector: the parent chooses which of several candidates is used ("book the one for $505", "the other card", "switch those items"). ALWAYS include.
- trigger: the parent only causes the child turn to happen. Include it when it is the request the child turn answers; otherwise omit.
- motivation: the parent states a task-level goal that constrains no value ("I need to update my address"). OMIT.
Before treating a parent as trigger or motivation, check that none of its content reaches the child in transformed form; if it does, it is a value parent.

STEP 1 -- Direct dependencies: every prior turn the target span directly depends on.
STEP 2 -- Recursive dependencies: for each dependency, identify what it depends on.
Continue until every path ends at a source node. No depth limit.

RULES

Tool calls:
- A tool result turn's ONLY direct dependency is the tool call turn that triggered it.
- A tool call turn's dependencies are the prior turns whose values were passed as arguments.
- The structure is a strict chain: tool_result -> tool_call -> [argument sources]. Never connect a tool result directly to its argument sources.

  CORRECT:
    "4": {"depends_on": [3]},     // tool result -> only its invoking tool call
    "3": {"depends_on": [2]}      // tool call -> turn that provided the argument

  WRONG:
    "4": {"depends_on": [3, 2]},  // tool result must NOT list argument sources directly
    "3": {"depends_on": [2]}

Graph structure:
- Each turn appears at most once in ground_truth_deps, keyed by its turn index as a string.
- A parent's index MUST be strictly smaller than its child's. Never cite a later turn.
- Every node must have a path to the target turn. Do not include nodes that nothing reaches.
- Source nodes are included with depends_on: [] and spans: {}.
- If the same turn is reachable by several paths, include it once.
\end{lstlisting}
\end{figure*}

\begin{figure*}[t]
\begin{lstlisting}[
  style=promptstyle,
  firstline=46,
  lastline=84,
  firstnumber=46,
  title={\textbf{Phase 2: Provenance Annotation Prompt} (part 2 of 3).}
]
Spans:
- Every edge has a span; every span belongs to an edge. A node's spans keys must be exactly its depends_on entries.
- A span is a SUBSTRING, not a summary. Before writing one, locate it in the parent turn as rendered below and copy the characters exactly, contiguously, in order. If a plain text search for your span would not find it in that turn, it is wrong.
- When the parent is a tool result, quote a contiguous slice of the RAW content, including whatever braces, quotes and field names it contains. Never reformat it into prose, a list, or a readable summary.
- Copy only characters that are present in the turn. Do not add quotation marks, brackets or wrappers around your span, and do not drop any that are there. Conversations differ in how they render tool calls -- match the turn in front of you, not the examples below.
- Never quote from the child turn, and never paraphrase. If you cannot quote the parent verbatim, the EDGE is wrong: fix the edge, not the quote.
- Keep each span under 120 characters. To cover more, add more spans -- do not stretch one.
- Several spans from one parent: use "11", "11#2", "11#3" as keys on the same node.
- Span width:
  * a composite value (an address, a product variant, a list of items) needs every attribute the target states -- use one span key per contiguous piece ("11", "11#2", ...), never one long rewritten span;
  * a correct comparative or superlative ("cheapest", "smallest") needs the comparison set, not only the winner -- again as several contiguous spans;
  * otherwise quote the minimum that carries the content.
- The target fields are GIVEN below. Copy name, target_turn_idx and target_text through unchanged; never reword the target, re-point it at another turn, or invent a new one.

Hallucinated targets:
- Wrong-value transformation (the span distorts an existing field or tool result) and stale-context error (the model follows an earlier task state despite later corrective evidence) are TRACEABLE: cite the turn holding the distorted field or the stale state.
- Unsupported invention (no prior turn supports the span) is UNTRACEABLE: give it no parent.
- Corrective evidence -- the later turn the model should have followed -- is NOT the source of a hallucinated span. Do not include it.
- Claims that restate agent policy (a refund window, a refund-method restriction, an inability to hold or reserve) are untraceable here: the domain policy is not part of the recorded conversation. Give them no value parent.

DEPENDENCY TYPE TAGS -- derive these from the graph you just built; they must agree with it.
Base (exactly one):
  direct  -- every parent of the target is a source node (one hop).
  chained -- at least one parent of the target has its own parents (two or more hops).
Structural (add all that apply):
  multi_source -- two or more distinct source nodes are ancestors of the target.
  hallucination -- the target span is factually wrong in context; the DAG traces where the wrong value came from, or has no parent if it was invented outright.

WORKED EXAMPLES -- the failures seen most often

Parent T8 is a tool result rendered as:
  [ 8] tool: {"results":[{"insurer_id":477,"insurer_name":"Blue Cross","out_of_network":false},
              {"insurer_id":851,"insurer_name":"Sun Life","out_of_network":false}]}

WRONG  "8": "Blue Cross (insurer ID: 477) - Sun Life (insurer ID: 851)"
       (reformatted; this string does not occur in T8)
WRONG  "8": "Blue Cross, Sun Life"
       (summary; not contiguous)
RIGHT  "8":    "{\"insurer_id\":477,\"insurer_name\":\"Blue Cross\""
\end{lstlisting}
\end{figure*}

\begin{figure*}[t]
\begin{lstlisting}[
  style=promptstyle,
  firstline=85,
  lastline=144,
  firstnumber=85,
  title={\textbf{Phase 2: Provenance Annotation Prompt} (part 3 of 3).}
]
       "8#2":  "{\"insurer_id\":851,\"insurer_name\":\"Sun Life\""
       (two contiguous raw-JSON slices, one per piece the target uses)

Tool calls are rendered differently in different conversations. Copy the form your
conversation uses.

Where the turn reads:
  [ 3] assistant: [CALL get_order_details({"order_id":"#W5166363"})]
RIGHT  "3": "[CALL get_order_details({\"order_id\":\"#W5166363\"})]"
       (or a contiguous slice such as "{\"order_id\":\"#W5166363\"}")

Where the turn reads:
  [11] assistant: RAG.query(query=['Aetna information'])
RIGHT  "11": "RAG.query(query=['Aetna information'])"
WRONG  "11": "[CALL RAG.query(query=['Aetna information'])]"
       (the [CALL ...] wrapper is not in this turn -- do not add it)

Where the turn reads:
  [15] tool: ["Wanting to run faster, yet needing to slow down seems contradictory."]
RIGHT  "15": "Wanting to run faster, yet needing to slow down"
WRONG  "15": "\"Wanting to run faster, yet needing to slow down\""
       (the extra quotation marks are not where you put them -- do not add them)

BEFORE YOU ANSWER, check each target against this list and fix anything that fails:
  1. name, target_turn_idx and target_text are exactly as given in TARGETS, and target_text appears verbatim in the target turn.
  2. The target turn is a node in ground_truth_deps.
  3. Every parent index is smaller than its child's.
  4. Every node has a path to the target.
  5. For every span, a plain text search of the parent turn as rendered would find it, character for character, with no wrapper or quotation marks you added. Tool-result spans are raw JSON, not prose. Each is under 120 characters.
  6. Every edge has a span and every span has an edge.
  7. Tool results depend on exactly one tool call turn.
  8. The tags match the graph: direct vs chained by hop count, multi_source only with two or more source ancestors.

CONVERSATION (turn index | role: content):
{conversation}

TARGETS:
{targets_json}

Return a single JSON object. No other text. Schema:
{
  "sim_id": "{sim_id}",
  "targets": [
    {
      "name": "<name>",
      "target_turn_idx": <int>,
      "target_text": "<exact text>",
      "comment": "<one sentence tracing the main dependency chain>",
      "dep_type_tags": ["<tag>", ...],
      "ground_truth_deps": {
        "<turn_idx>": {
          "depends_on": [<turn_idx>, ...],
          "spans": {
            "<turn_idx>": "<exact span from that turn>"
          }
        }
      }
    }
  ]
}
\end{lstlisting}
\end{figure*}

\subsection{ConFETTI Dataset Corrections}
\label{sec:appendix-confetti-corrections}

The ConFETTI dataset~\citep{alkhouli2025confetti} provides 109 multi-turn conversations (506 function-calling turns) used in the ConFETTI split of \textsc{MTCABench}.
Prior to annotation, we applied input disambiguation corrections following a previously described LLM-assisted human-in-the-loop methodology~\citep{agarwal2026switchcraft}.

86 turns (17\%) referenced dates without specifying a year, while the gold labels contained hardcoded year values.
Without the year present in the conversation, the correct date is unattributable to any conversational turn---the model must either rely on world knowledge or guess---making the attribution target ill-defined.
We appended the missing year (or month and year) to the relevant user utterances so that the conversation unambiguously determines the expected function call.
These changes do not alter gold labels.

\subsection{Annotation Quality and Adjudication}
\label{sec:appendix-annotation-quality}

Because the gold provenance DAGs are themselves LLM-produced, their quality
bounds every number in this paper. Phase~3 therefore validates them and
selects the final gold DAGs in five steps, 3a--3e, reported separately below.

\paragraph{Step 3a: the well-formedness sweep.}
A provenance DAG can be decidably checked for well-formedness from the conversation and the
annotation alone. We run the following checks on all 3,984 targets from GPT-5.4's annotations:

\begin{itemize}\itemsep2pt
  \item \textbf{Tags contradict the graph} (242). The dependency-type tag asserts
  a shape the graph does not have: tagged \textit{direct} while some parent has
  parents of its own, tagged \textit{chained} while every parent is already a
  source, or tagged \textit{multi\_source} with fewer than two source nodes.
  No released target can carry this defect, since step~3e derives the structural 
  tags from the graph directly instead. 

  \item \textbf{Span not verbatim in the turn it cites} (208). The quoted
  span does not occur verbatim in the parent turn. Most cases are a dropped
  character,punctuation error, minor formatting issue, or a paraphrased quote rather than a wrong turn.
  We treat these cases as well-formedness defects rather than correctness defects.

  \item \textbf{Span quoted from the child turn} (139). A special case of the
  above, separated because it is a distinct failure mode: the annotator quoted
  the text from the turn being explained instead of from its source, which makes
  the edge circular as evidence.

  \item \textbf{Node with no path to the target} (128). The graph declares a turn
  as a node but is not reachable from the target depends on it, so it cannot
  participate in the target's provenance. An unreachable node is a false positive against
  every method and breaks the assumption that all nodes contribute to the target's provenance.

  \item \textbf{Parent indexed later than its child} (50). An edge runs forward
  in time; a turn cannot draw on a turn the model had not yet seen.

  \item \textbf{Target text absent from its own turn} (41). If the span being
  explained does not occur verbatim in the assistant turn, there is nothing to
  attribute.

  \item \textbf{Edge without a span} (26). An asserted dependency carries no
  quoted evidence, so the claim cannot be checked or scored at span level.

  \item \textbf{Malformed tool-call chain} (15). A tool-result turn depends on
  something other than exactly one tool-call turn, violating the deterministic 
  structural edge the annotation protocol requires.
\end{itemize}

Of the 3,984 targets, 751 (18.9\%) involve at least one of these defects, leaving
3,233 (81.1\%) free of them.

\paragraph{Step 3b: independent human review.}
Three authors independently judged \texttt{GPT-5.4}'s graphs for the same 100
stratified targets (50 per split, balanced across dependency types) as
\textit{Correct}, \textit{Minor issues}, or \textit{Wrong}. Acceptance (correct or minor) is 95\%.
Authors had three-way exact agreement on 64 of 100: reviewers rarely disagreed about \emph{whether} a DAG was defective, only about severity. 
Of the 36 disagreements, 20 are Correct/Minor splits and only 5 are three-way. 
Adjudicating all 36 produced the consensus labels, and authors reached near-perfect agreement: 
pairwise Cohen's $\kappa$ of 1.00, 0.96, and 0.96 across the three annotator pairs, for a mean pairwise $\kappa$ of 0.97, with unanimous agreement on 98/100 items.
The refined prompt with the adjudication guidelines was used in step~3c.

\paragraph{Step 3c: cross-annotation validation.}
We sweep every phase-2 annotation for well-formedness and score the four model 
annotations against one another (Table~\ref{tab:reannotation}).
\texttt{GPT-5.4} was annotated first, under an earlier version of the phase-2 prompt
that lacked the span-construction rules described in Appendix~\ref{sec:appendix-prompts}.
The other three---\texttt{GPT-5.6-Sol} and \texttt{GPT-6-Luna} at medium reasoning effort, 
\texttt{GPT-6.1-Sol} at high---share the prompt in Listing~\ref{lst:prompt-phase2} and 
independently regenerate the DAG for each target.

The three re-annotation models agree with one another at 0.911--0.926 node $F_1$ and with
\texttt{GPT-5.4} at 0.851--0.861 (Table~\ref{tab:reannotation}). Part of that gap
is attributable to the prompt rather than to the model. Empty node sets---both annotations agreeing
a target is unsupported---count as agreement. Within the three that share a prompt,
higher reasoning effort bought a lower defect rate but no higher agreement.

\begin{table}[h]
\centering
\small
\begin{tabular}{lrr}
\toprule
Annotator & Defect rate & Node $F_1$ vs.\ others \\
\midrule
GPT-5.6-Sol            &  1.7\% & 0.913--0.926 \\
GPT-6.1-Sol (high)     &  1.5\% & 0.911--0.926 \\
GPT-6-Luna             & 12.9\% & 0.911--0.913 \\
GPT-5.4 (earlier prompt) & 18.9\% & 0.851--0.861 \\
\bottomrule
\end{tabular}
\caption{\textbf{Annotator identity drives well-formedness, not structure.}
Defect rate is the step-3a sweep applied to each annotation's graphs against the
targets as phase~1 selected them. Node $F_1$ is the range of pairwise agreement with
the other annotations.}
\label{tab:reannotation}
\end{table}

\paragraph{Step 3d: blind adjudication of the residual.}
We define agreement between two annotations on a target when their node sets reach $F_1 \geq 0.8$. 
Over the 3,960 targets all three re-annotations cover, a single
pair of annotators leaves 328 targets (8.3\%) unresolved, while across all three
at least one pair agrees on 3,916 (98.9\%). The residual is \textbf{44 targets (1.1\%)} on 
which no pair agrees. Those 44 targets, along with 8 near-threshold cases, were adjudicated blind by one human author. The four phase-2 candidate DAGs were
presented as A--D, shuffled per target, with the letter-to-annotator mapping
recorded but not shown. 

The adjudicator chose a single best graph for 45 targets, judged 3 to have
several equally good candidates, and proposed a modification on 4. Ties and modifications were resolved by written criteria.
One target was dropped, its target text being absent from the turn it was attributed to. No annotator
dominates: of the 51 graphs adopted, 19 come from GPT-5.6-Sol, 14 from
GPT-6.1-Sol, 10 from GPT-5.4 and 7 from GPT-6-Luna.

\paragraph{Step 3e: assembling the released gold.}
From steps 3a--3d, each target has up to four candidate DAGs. We select them 
by majority vote:

\textbf{Majority pool.} For each candidate DAG we take its node set---every turn
appearing as a node or as a parent, with compound span keys collapsed to their
turn. A candidate joins the pool when its node set is \emph{exactly} identical to
at least one other candidate's. 
forms.

\textbf{Selection.} Candidates are ordered pool members first, then non-members,
each group in a fixed annotator order, and we take the first that passes the
step-3a sweep; failing that the original annotation if it passes.
The vote therefore \emph{selects} one of the graphs wholly, it does not merge graphs or
prune nodes outside the agreed set. For the 52 adjudicated targets the
adjudicated graph is used directly and no vote is taken.


\begin{table}[t]
\centering
\small
\begin{tabular}{llrr}
\toprule
Gold & DAG source & $n$ & Defective \\
\midrule
Original    & GPT-5.4 throughout          & 3,984 & 18.9\% \\
Adjudicated & original $+$ 52 adj. & 3,973 & 17.7\% \\
Final    & majority vote, above        & 3,970 & \textbf{0.0\%} \\
\bottomrule
\end{tabular}
\caption{\textbf{Gold versions.} Replacing only the adjudicated targets barely
moves the defect rate; majority vote over the re-annotations removes the defects entirely.}
\label{tab:gold-versions}
\end{table}

\textbf{Final well-formedness sweep.}
Structural tags \textit{direct}/\textit{chained} and \textit{multi\_source} are
functions of the graph and are derived from it.
\textit{hallucination} and \textit{tool\_call} are judgements no graph determines
and are carried over unchanged.

This yields \textbf{3,970 targets with no well-formedness defect}, against 18.9\%
originally (Table~\ref{tab:gold-versions}).
 
\clearpage
\section{Tokengeist Procedure Walkthrough}
\label{sec:appendix-procedure}

\begin{figure*}[tp]
  \centering
  \includegraphics[width=\textwidth]{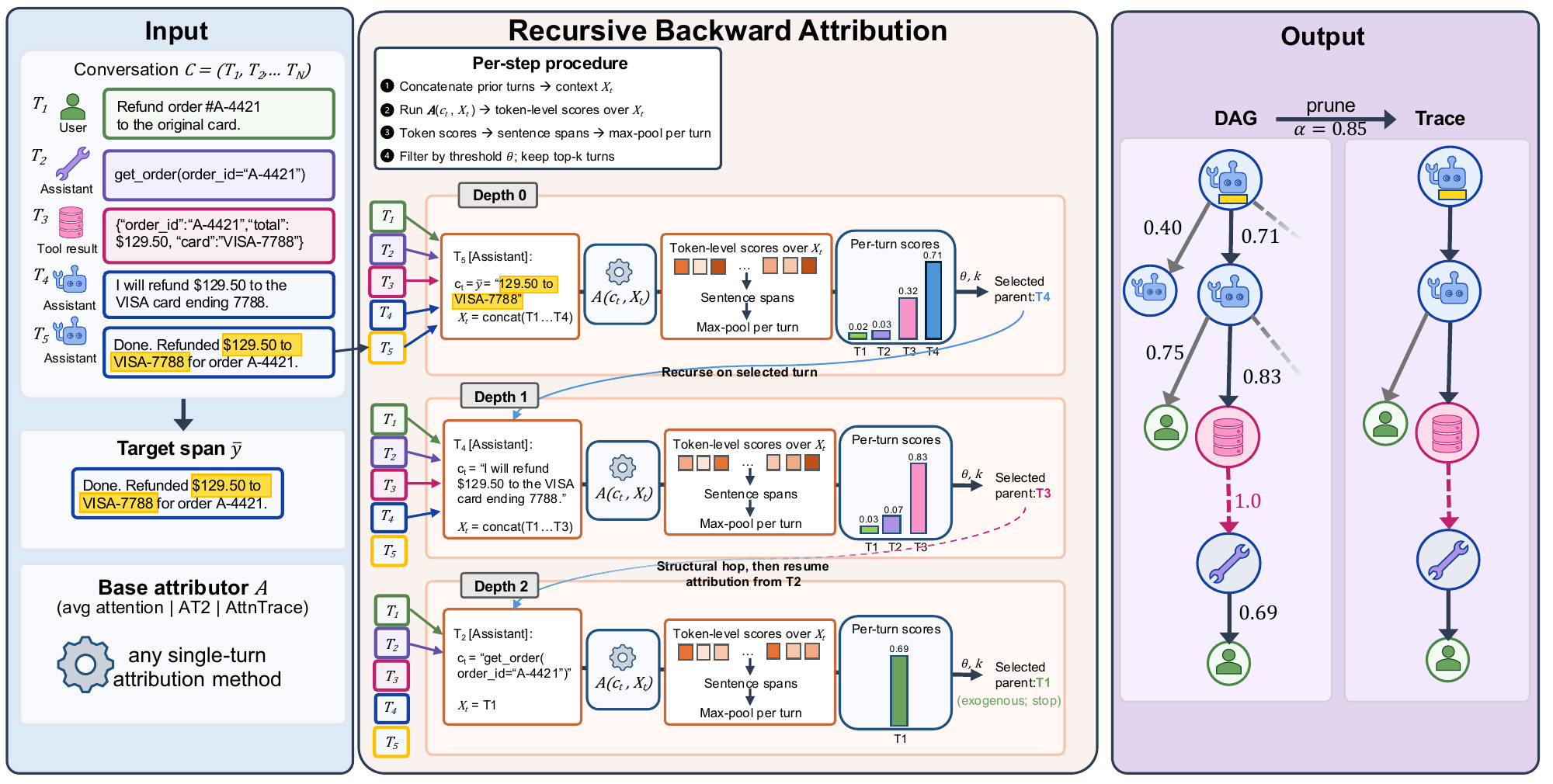}
  \caption{\textbf{Tokengeist procedure on a five-turn refund-tool example.}
  Stage~1 lists the inputs (the conversation, the target span $\bar{y}$, and the base attributor $\mathcal{A}$).
  Stage~2 unrolls three recursion steps and shows the per-step substeps (concatenate, score, map to turns, top-$k$) once, in a legend box that applies to every step.
  Stage~3 shows the constructed provenance DAG and the final trace after relative-threshold extraction with $\alpha\!=\!0.85$ (Section~\ref{sec:tokengeist}).
  Solid arrows denote attribution edges; the dashed magenta arrow denotes the deterministic structural edge that bridges a tool-result turn to its invoking tool-call turn.}
  \label{fig:procedure-walkthrough}
\end{figure*}

Figure~\ref{fig:procedure-walkthrough} visualizes the \textsc{Tokengeist} procedure formalized in Section~\ref{sec:tokengeist} on a concrete five-turn tool-using example.
It is intended as a reference for readers following the algorithm step-by-step; the body of the paper relies on the compact teaser in Figure~\ref{fig:method}.

\paragraph{Why this example.}
The five-turn refund conversation is constructed to exercise all three role-aware dispatch branches of the recursion in a single backward trace: a recurse-into-assistant hop ($T_5\!\to\!T_4$), a tool-result structural hop ($T_3\!\to\!T_2$, dashed magenta), and termination at an exogenous user leaf ($T_1$).
Real \textsc{MTCABench} traces typically combine several such branches; the depth and branching statistics across the full benchmark are reported in Figure~\ref{fig:depth_hist}.

\paragraph{What the example does not exhibit.}
Two aspects of the algorithm are implicit in the figure but not visually exercised by this single-chain trace.
First, \textbf{DAG deduplication}: when a turn is reached via a second attribution path, a convergent edge is added to the existing node rather than creating a duplicate (Section~\ref{sec:tokengeist}).
Second, \textbf{multi-source branching}: the bar chart in each Stage~2 box collapses to a single dominant turn here, whereas multi-source targets retain the top $k\!=\!3$ candidates and recurse into each.

\paragraph{Reading the figure alongside the algorithm.}
The Stage~2 legend, the role-aware dispatch arrows, and the stopping criteria all correspond one-to-one with the per-step procedure in Section~\ref{sec:tokengeist} and its reference implementation in \texttt{RecursiveProvenanceTracer.\allowbreak\_trace\_recursive}.
The pruning rule shown in Stage~3 is the relative-threshold extraction defined in Equation~\ref{eq:rel-threshold}; the choice $\alpha\!=\!0.85$ is justified by the sweep in Appendix~\ref{sec:appendix-experiments}.

Figure~\ref{fig:intuition} provides a complementary conceptual view, contrasting single-pass and recursive attribution at the level of context-window coverage.

\begin{figure*}[tp]
  \centering
  \includegraphics[width=\textwidth]{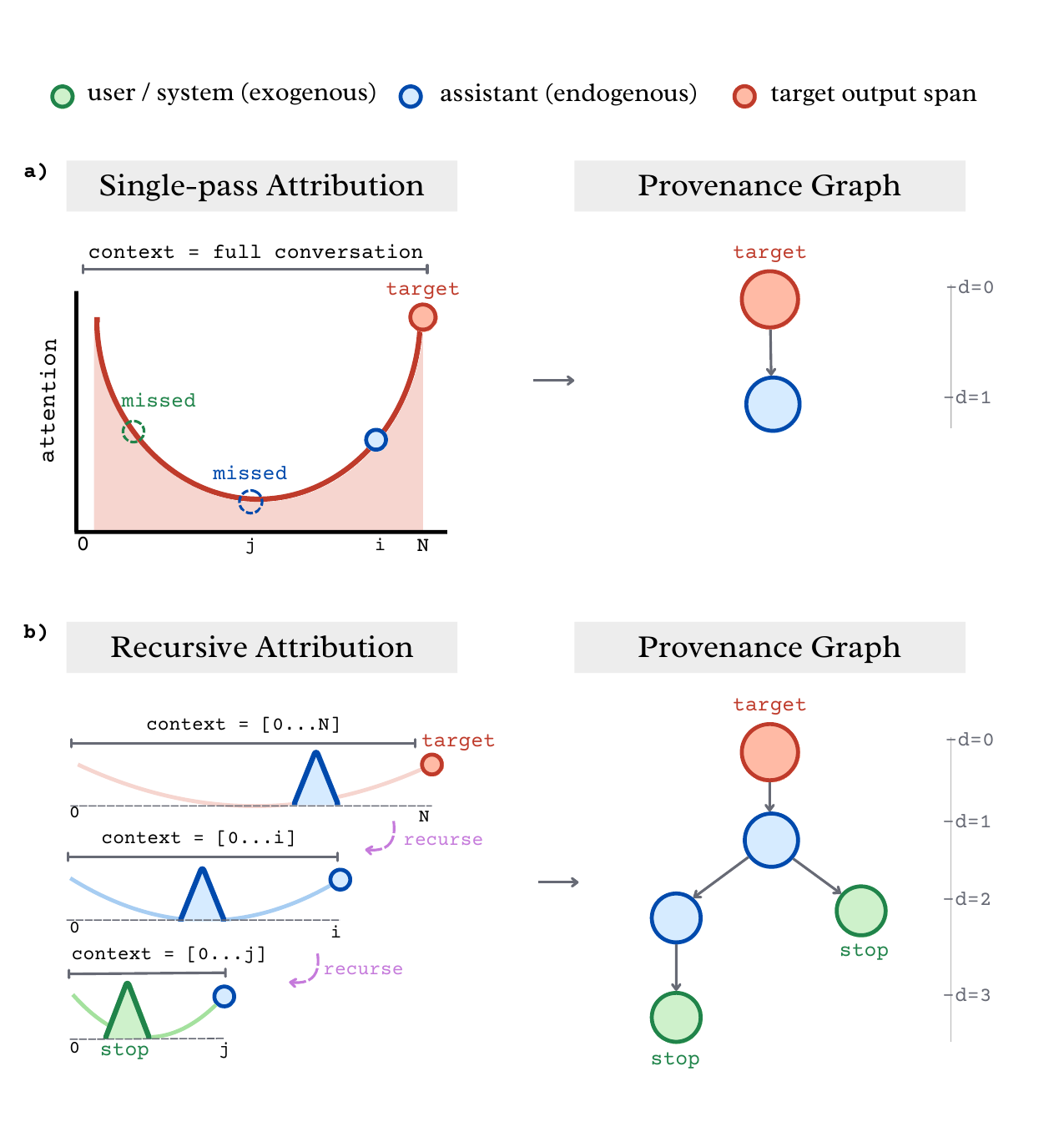}
  \caption{\textbf{Provenance collapse under flat attribution and recursive recovery.}
  \textit{Top:} Single-pass (flat) attribution computes attention from the target output span over the entire conversation in a single step. The intermediate assistant turns have already absorbed the content of the sources behind them, so the target has little reason to attend past those turns and the sources themselves (blue) are missed or underweighted; positional degradation over long contexts~\citep{liu-etal-2024-lost} compounds this but is not its cause (Appendix~\ref{sec:appendix-signal-depth}). The resulting provenance graph has depth~1 and loses the multi-hop structure.
  \textit{Bottom:} Recursive attribution traces backward through intermediate assistant turns, re-running attribution on each one's context window until exogenous turns (green) are reached. The first call sees the full context $[0\ldots N]$ but each recursive call starts from the position of the newly found intermediate, so the effective context window shrinks at every hop. This recovers the full depth-$d$ provenance DAG, correctly attributing the target to its true sources.}
  \label{fig:intuition}
\end{figure*}

\section{Hallucination Provenance Analysis}
\label{sec:appendix-hallucination}

For hallucination-tagged targets, the primary provenance question is pinpointing the context that led the model to produce the questionable span. We call these turns \textit{hallucinated sources}.
They may contain a stale value, a distorted field, an overridden instruction, or other context that produced or propagated the erroneous span.

\paragraph{Hallucination taxonomy.}
  We observe three recurring patterns:
  \begin{enumerate}
    \item \textit{Wrong-value transformations} (traceable): the span distorts an existing field or tool result.
    \item \textit{Stale-context errors} (traceable): the model follows an earlier task state despite later corrective evidence.
    \item \textit{Unsupported inventions} (untraceable): no prior turn supports the generated span, so there is no source to recover.
  \end{enumerate}

\paragraph{Hallucinated and corrected targets.}
Attribution follows the target span it is asked to explain.
For the original hallucinated targets, the task is to trace where the questionable generated span came from.
This is useful diagnostically: the trace identifies where an erroneous value entered or propagated through the agentic workflow.
To test the complementary setting, we also construct a corrected ConFETTI subset by replacing 19 wrong-value hallucinations with reviewed corrected values, preserving the original conversation and upstream provenance DAG.

For example, one ConFETTI case changes a hallucinated appointment time, "from 1:35 AM to 02:00 PM," to the corrected span, "from 1:35 AM to 02:00 AM," where the corrected target depends directly on an earlier tool result containing the true end time of 2:00 AM. The full DAG traces that evidence back through the tool call and the user's request.

\begin{table*}[t]
\centering
\small
\setlength{\tabcolsep}{4pt}
\begin{tabular}{lccc|ccc}
\toprule
& \multicolumn{3}{c|}{Hallucinated targets}
& \multicolumn{3}{c}{Corrected targets} \\
\cmidrule(lr){2-4}\cmidrule(lr){5-7}
Method & Halluc. F1 & Corr. F1 & Overall F1 & Evidence F1 & Node F1 & Source F1 \\
\midrule
\textsc{Tg} + Attention & 0.657 & 0.281 & \textbf{0.616} & \textbf{0.903} & \textbf{0.792} & \textbf{0.484} \\
\textsc{Tg} + AT2       & \textbf{0.664} & 0.263 & 0.612 & 0.877 & 0.751 & 0.453 \\
\textsc{Tg} + AttnTrace & 0.636 & 0.280 & 0.598 & 0.272 & 0.574 & 0.423 \\
Flat Attention          & 0.155 & \textbf{0.296} & 0.322 & \textbf{0.903} & 0.375 & 0.132 \\
Flat AT2                & 0.178 & 0.286 & 0.333 & 0.877 & 0.363 & 0.184 \\
Flat AttnTrace          & 0.249 & 0.207 & 0.316 & 0.237 & 0.425 & 0.323 \\
\bottomrule
\end{tabular}
\caption{\textbf{Hallucinated-target versus corrected-target attribution.}
Hallucinated-target scores evaluate attribution on the originally generated erroneous span.
Halluc. F1 measures recovery of the source of the wrong-value span; Corr. F1 is auxiliary and asks whether the same attribution also recovers evidence for the corrected answer.
Corrected-target scores evaluate attribution after replacing ConFETTI wrong-value hallucinations with corrected spans.}
\label{tab:hallucination_corrected_attribution}
\end{table*}

Table~\ref{tab:hallucination_corrected_attribution} shows both sides of this distinction.
On hallucinated targets, \textsc{Tokengeist}'s recursive tracing improves hallucinated-source F1 by 40--50 points over flat attention and AT2, while corrective-source F1 remains low because the corrective evidence is not the source of the hallucinated span.
On corrected targets, flat attention and AT2 recover direct evidence well, but \textsc{Tg}+Attention improves Node F1 from 0.375 to 0.792 and Source F1 from 0.132 to 0.484.

\begin{figure}[htp]
  \centering
  \includegraphics[width=\linewidth]{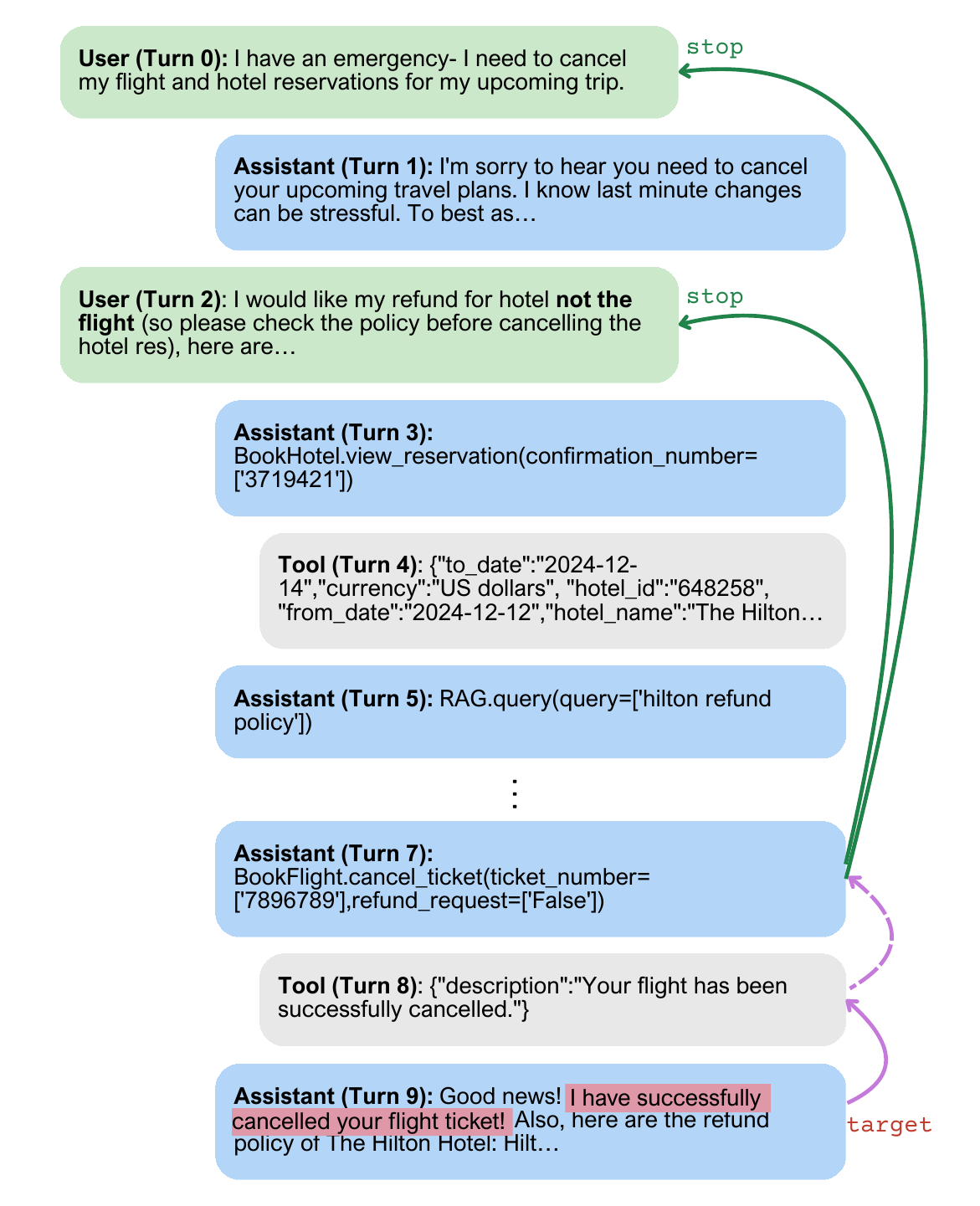}
  \caption{\textbf{Hallucination example from ConFETTI.}
  The gold provenance trace for the target span ``I have successfully cancelled
  your flight ticket'' (Turn~9) is shown with arrows indicating directed
  dependencies.}
  \label{fig:hallucination_example}
\end{figure}

\paragraph{Provenance example.}
Figure~\ref{fig:hallucination_example} illustrates the annotation distinction in an ignored-correction case.
The user requests flight and hotel cancellations (Turn~0), then immediately corrects to hotel only---``not the flight''---in Turn~2.
The agent cancels the flight anyway and reports success in Turn~9.
The hallucinated provenance points to the stale cancellation intent the agent followed, while the corrective evidence points to the later instruction it should have followed.
This is the ideal analysis target: a graph that separates the context that produced the wrong action from the evidence that would support the corrected action.
As Table~\ref{tab:hallucination_corrected_attribution} shows, the former is the attribution task \textsc{Tokengeist} is designed for, while the latter is harder to recover from a hallucinated target.

\section{Attribution with the Generating Model}
\label{sec:appendix-selfgen}

In the main experiments the attribution model reads conversations another source produced---human-simulated dialogs in ConFETTI, \texttt{gpt-4.1} agents in TauBench---so its attention may describe a post-hoc reading of the transcript rather than how the text was produced.
We therefore replay a subset of \textsc{MTCABench} so that every assistant turn comes from the attributing model (the \emph{matched-generator} condition) and rerun the main comparison.
Every main-paper finding holds: recursive attribution outperforms flat attribution with the same backend in all four replays, by a margin indistinguishable from the one on the original transcripts once provenance-graph size is controlled, and on self-generated turns the attention \textsc{Tokengeist} reads is numerically the attention that generated the text.

\subsection{Replay protocol}
\label{sec:appendix-selfgen-protocol}

Each conversation is replayed turn by turn: user and tool-result turns are copied from the record, and every assistant turn is regenerated by the attribution model with the dialog's tool schemas in the system prompt (on TauBench, together with the agent instruction and domain policy).
Whether a turn is a tool call or a freeform reply is replayed from the record; the choice of function, arguments and wording is the model's.

A recorded tool result stays a valid continuation only if the regenerated call is the same call, so each tool-call turn draws up to eight candidates (the first greedy, the rest at temperature 0.8) and keeps the first one the BFCL AST checker~\citep{agarwal2026switchcraft} accepts against the recorded call---up to argument order, value types, list wrapping and paraphrased multi-word string arguments (embedding similarity 0.85).
The two calls below, for example, are accepted as the same call: the gold answer wraps each argument's acceptable values in a list, and the model emits plain scalars in a different order.

\noindent\begin{minipage}{\columnwidth}
\begin{lstlisting}[style=promptstyle,frame=none,xleftmargin=0pt,xrightmargin=0pt,breakindent=1.5em,aboveskip=0.4em,belowskip=0.4em]
Telecom.SendVerificationToken(to=["csmith@gmail.com"], channel=["EMAIL"], last_name=["Smith"])

Telecom.SendVerificationToken(to="csmith@gmail.com", last_name="Smith", channel="EMAIL")
\end{lstlisting}
\end{minipage}

\noindent A multi-word string argument need only be close: \texttt{query=["tips to run faster"]} also accepts \texttt{query="ways to boost running speed"}.

Reply turns draw four candidates and keep the one closest to the record under MiniLM similarity; on TauBench, where the simulated user responds to the agent's commitments, a reply below 0.50 falls back to the recorded one (1.5--1.7\% of replies).
If nothing passes, the recorded turn is kept and marked as not model-generated; when every candidate has the wrong type---typically a tool call where the record asks the user for a missing argument---a one-line instruction fixing the turn type is appended for that turn only (16.4--20.2\% of assistant turns), which we call \emph{mode-forced}.\footnote{One of four fixed strings, one per benchmark and turn type; for a ConFETTI tool call it is \texttt{Call exactly one API on this turn. Emit only the call.}}

We replay all 109 ConFETTI dialogs and the 129 labeled TauBench airline simulations with Llama-3.1-8B and Qwen2.5-7B.
The regenerated transcripts are relabeled with the annotation protocol of Section~\ref{sec:annotation_protocol}, and a target is \emph{admitted} only if every turn on its gold provenance path was generated by the attribution model.
Attribution reads each conversation in the format the model generated it in---the chat template with the generation system prompt, which stays in the context but is never selected as a source---so the generation context of every turn except the mode-forced ones is reproduced exactly.

\subsection{Results}
\label{sec:appendix-selfgen-results}

\begin{table*}[t]
\centering
\small
\setlength{\tabcolsep}{3.5pt}
\begin{tabular}{lrrcccc|rrcccc}
\toprule
& \multicolumn{6}{c|}{\textit{ConFETTI}} & \multicolumn{6}{c}{\textit{TauBench}} \\
\cmidrule(lr){2-7}\cmidrule(lr){8-13}
Method & Convs & Targ. & Edge F1 & Node F1 & Src.\ Rec. & Span F1 & Convs & Targ. & Edge F1 & Node F1 & Src.\ Rec. & Span F1 \\
\midrule
\multicolumn{13}{l}{\textit{Llama-3.1-8B-Instruct}} \\
\midrule
Flat Attention          & 83 & 291 &           52.5 &           57.2 &           70.1 &           14.5 & 116 & 398 &           32.1 &           38.5 &           31.3 &            5.1 \\
Flat AT2                &    &     &           50.4 &           55.4 &           65.9 &           13.8 &     &     &           26.8 &           35.2 &           32.8 &            4.9 \\
Flat AttnTrace          &    &     &           39.4 &           60.9 &           80.7 &           22.8 &     &     &           18.0 &           30.0 &           46.9 &            6.7 \\
\midrule
\textsc{Tg} + Attention &    &     &  \textbf{83.0} &  \textbf{88.4} &  \textbf{97.5} &  \textbf{47.5} &     &     &  \textbf{58.7} &  \textbf{69.9} &           89.3 &  \textbf{20.4} \\
\textsc{Tg} + AT2       &    &     &           77.6 &           84.0 &           96.2 &           44.4 &     &     &           46.7 &           60.5 &           89.3 &           17.3 \\
\textsc{Tg} + AttnTrace &    &     &           60.8 &           76.2 &           96.8 &           41.9 &     &     &           25.2 &           38.7 &  \textbf{89.7} &           11.0 \\
\midrule
\multicolumn{13}{l}{\textit{Qwen2.5-7B-Instruct}} \\
\midrule
Flat Attention          & 77 & 242 &           49.6 &           53.3 &           63.9 &           12.9 &  97 & 319 &           34.7 &           37.6 &           27.0 &            4.1 \\
Flat AT2                &    &     &           48.5 &           53.4 &           69.0 &           13.2 &     &     &           32.4 &           38.1 &           30.0 &            4.5 \\
Flat AttnTrace          &    &     &           38.0 &           55.9 &           72.1 &           20.4 &     &     &           19.0 &           31.7 &           37.0 &            5.4 \\
\midrule
\textsc{Tg} + Attention &    &     &  \textbf{79.8} &  \textbf{86.0} &           93.7 &  \textbf{46.5} &     &     &  \textbf{65.0} &           72.2 &  \textbf{90.8} &  \textbf{15.7} \\
\textsc{Tg} + AT2       &    &     &           78.0 &           85.0 &           93.0 &           44.9 &     &     &           62.5 &  \textbf{72.3} &           90.6 &           15.2 \\
\textsc{Tg} + AttnTrace &    &     &           59.1 &           73.1 &  \textbf{96.0} &           40.8 &     &     &           33.5 &           51.1 &           90.7 &           12.2 \\
\bottomrule
\end{tabular}
\caption{\textbf{Attribution under a matched generator.} Each model attributes the transcripts it generated itself, scored over the admitted targets of the conversations it reproduced (Convs, Targ.). All scores are percentages; \textsc{Tg} = \textsc{Tokengeist}; bold marks the best recursive method per metric per model. Recursive attribution beats its flat counterpart in every cell.}
\label{tab:selfgen_selfcross}
\end{table*}

\paragraph{Yield.}
Between 44.1\% and 50.8\% of tool calls are reproduced; most failures are a different tool on ConFETTI (60.8--67.7\%) and a different number of calls on TauBench (47.1--57.6\%), where \texttt{gpt-4.1} often issues calls in parallel.
Admission leaves 242--291 ConFETTI and 319--398 TauBench targets per model (Table~\ref{tab:selfgen_selfcross}).
The admitted set is limited as much by the sampling budget as by the model: a tool-call turn draws at most eight candidates and a reply four, all but the first at a single temperature, and a turn that fails is dropped rather than retried, so a wider candidate pool under the same acceptance criterion would admit more turns.

\paragraph{The main comparison is unchanged.}
In all four replays and for every backend, \textsc{Tokengeist} outperforms flat attribution by the margins reported in the main results (Table~\ref{tab:selfgen_selfcross}): on its own ConFETTI transcripts Llama-3.1-8B reaches 83.0 Edge F1 with \textsc{Tg}~+~Attention against 52.5 flat, and 58.7 against 32.1 on TauBench.
Scores are also usually higher on the model's own transcripts than on the original ones over the same conversations, and \textsc{Tg}~+~Attention is higher in every replay: 88.4 Node F1 against 81.7 on ConFETTI and 69.9 against 62.5 on TauBench, with the 95\% conversation-bootstrap interval for the difference excluding zero in both.
One likely reason is that on its own transcripts the attention \textsc{Tokengeist} reads is the attention that produced the text, as we verify below, whereas on the originals it is a reading of text the model did not write.
The two conditions score the same conversations but not the same targets: admission selects the shorter gold provenance chains, with mean gold depth 2.7 and 2.8 among admitted ConFETTI targets against 3.3 and 3.6 over all targets of the same regenerated transcripts, and 3.4 and 3.7 against 4.8 and 5.0 on TauBench (Llama-3.1-8B and Qwen2.5-7B).
The comparison between the two conditions is therefore indicative rather than controlled.


\paragraph{Attribution reads the generation-time attention.}
Our experiments attribute completed transcripts, so \textsc{Tokengeist} reads attention from a single teacher-forced pass, whereas the text was produced token by token with a key--value cache.
For unforced self-generated target turns we recompute the generation-time attention---a prefill plus one cached decoding step per target token---and compare the two attention tensors over layers, heads, target tokens and context tokens.
They agree to numerical precision on all 24 TauBench and 12 ConFETTI turns (Llama-3.1-8B).
The mean absolute difference over all entries of the tensor is below $10^{-6}$ in the median turn, and no single weight anywhere differs by more than $0.07$; collapsing each tensor to one weight per context token (the mean over layers, heads and target tokens) leaves two profiles that correlate at Pearson $r > 0.9999997$ in every turn; and summing those profiles over each source turn gives an identical ranking (same top three, Spearman $\rho = 1$).
Without the generation system prompt the same turn reproduces the top three for only 0.0\% and 8.3\% of turns (median $\rho$ 0.68 and 0.64), so the agreement comes from reproducing the generation context, not from any reading of the transcript.

\end{document}